\pdfoutput=1   %
\documentclass{article}

\usepackage[preprint]{neurips_2026}   %

\usepackage[utf8]{inputenc} %
\usepackage[T1]{fontenc}    %
\usepackage{url}            %
\usepackage{booktabs}       %
\usepackage{amsfonts}       %
\usepackage{nicefrac}       %
\usepackage{microtype}      %
\usepackage[table]{xcolor}  %

\usepackage{algorithm}
\usepackage{algpseudocode}
\usepackage{amsmath}
\usepackage{graphicx}
\usepackage{todonotes}
\usepackage{makecell}
\usepackage{amssymb}
\usepackage{multirow}
\usepackage{booktabs}
\usepackage{makecell}
\usepackage{enumitem}
\usepackage{wrapfig}
\definecolor{colorfuture}{HTML}{FF0000}
\definecolor{colorcurrent}{HTML}{00B0F0}
\usepackage[pagebackref=false, breaklinks=true, colorlinks, citecolor=citecolor, linkcolor=linkcolor, bookmarks=false]{hyperref}
\definecolor{citecolor}{HTML}{0071BC}
\definecolor{linkcolor}{HTML}{ED1C24}

\title{Video-Mirai: Autoregressive Video Diffusion Models Need Foresight}

\author{%
  Yonghao Yu$^{1}$ \quad
  Lang Huang$^{2,3}$\thanks{Corresponding author.} \quad
   Runyi Li$^{1}$ \quad
  Zerun Wang$^{1}$ \quad
  Toshihiko Yamasaki$^{1}$
  \\[0.3em]
  $^{1}$The University of Tokyo \quad
  $^{2}$National Institute of Informatics \quad
  $^{3}$SOKENDAI
  \\[0.3em]
{\normalfont\fontsize{9.pt}{10.5pt}\selectfont
\texttt{\{y\_yu, r\_li, ze\_wang, yamasaki\}@cvm.t.u-tokyo.ac.jp, lang@nii.ac.jp}}
}
\makeatletter
\renewcommand{\@noticestring}{}
\makeatother

\begin{document}
\maketitle
\vspace{-0.25in}
\begin{abstract}
Causal video generators must predict from the past, but they need not learn only from it. In streaming autoregressive video diffusion, each emitted segment becomes a commitment that future segments must preserve. Standard training, however, does not explicitly supervise causal states with future-informed representation targets. This creates what we call a representation-level planning gap: states that fit the current segment may discard identity, layout, and motion information needed for a consistent future. We introduce Video-Mirai, a training-only method that closes this gap without changing causal inference: the generator rolls out causally, a frozen foresight encoder reads the completed rollout non-causally, and a lightweight predictor distills the resulting stopped-gradient targets into causal states. Future frames supervise representations, never generator inputs. At inference, the encoder and predictor are discarded, leaving the original architecture, per-step FLOPs, and KV-cache behavior unchanged. Video-Mirai improves a strong Causal-Forcing baseline on 5-second VBench from 83.8 to 84.6 in terms of Total Score. On 30-second rollouts beyond the training horizon, subject consistency improves from 84.9 to 88.5 and background consistency from 90.2 to 91.9. Ablations identify future-conditioned targets as a key ingredient, and probes show that future frames become more decodable from current features. Causality should constrain inference, not representation supervision.
Our study highlights that visual autoregressive models need foresight. Project Page: \url{https://y0uroy.github.io/Video-Mirai}.
\end{abstract}
\begin{figure*}[h]
  \centering
    \vspace{-6.5mm}
    \includegraphics[width=0.92\linewidth]{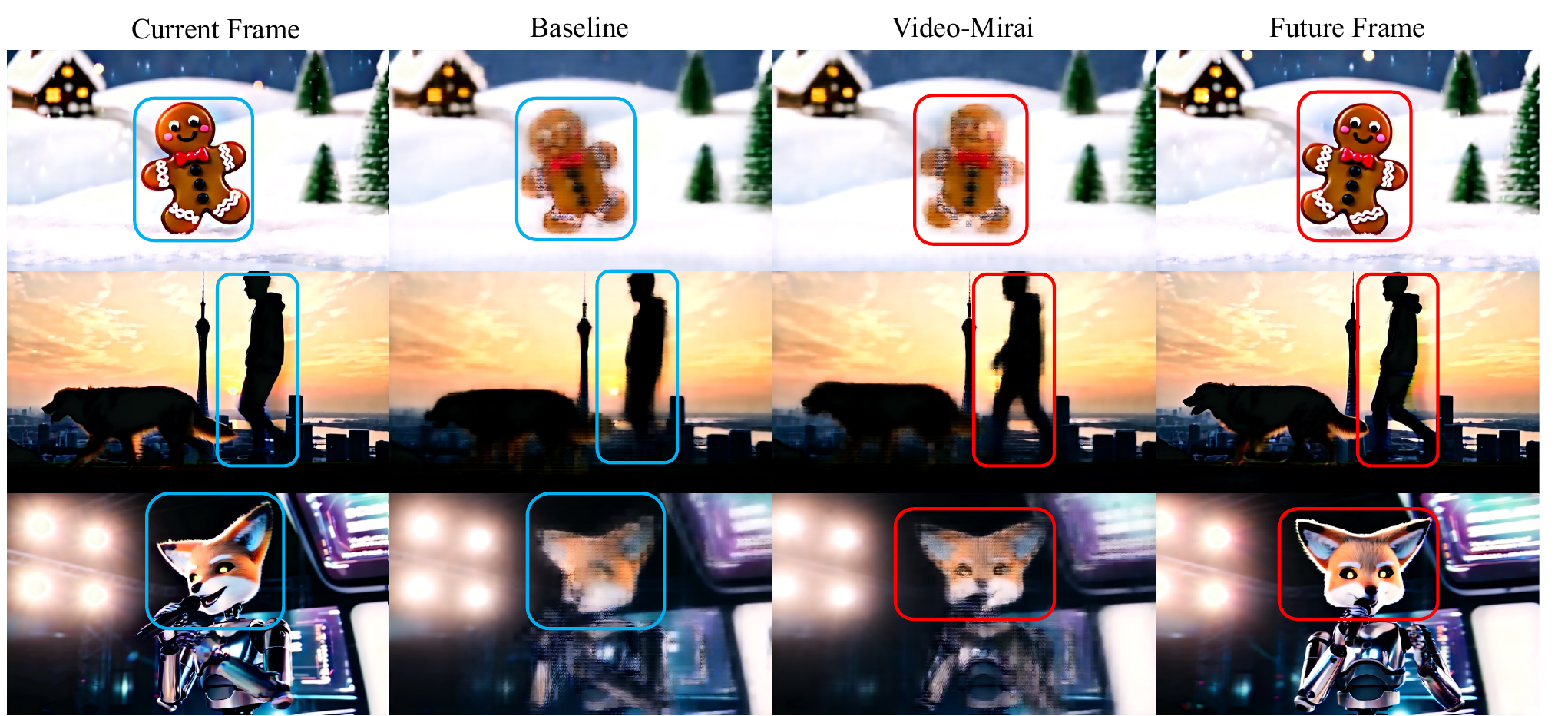}
    \vspace{-2mm}
    \caption{\textbf{Video-Mirai makes the future more decodable.} An MLP readout reconstructs future RGB from the frozen causal generator’s current hidden state. Left to right: current frame, baseline readout, Video-Mirai readout (ours), future frame. \textcolor{colorcurrent}{blue}/\textcolor{colorfuture}{red}: regions matching the \textcolor{colorcurrent}{current}/\textcolor{colorfuture}{future}.}
\label{fig:visualization}

\end{figure*}

\section{Introduction}
\label{sec:introduction}
\begin{figure*}[t]
  \centering
    \includegraphics[width=\linewidth]{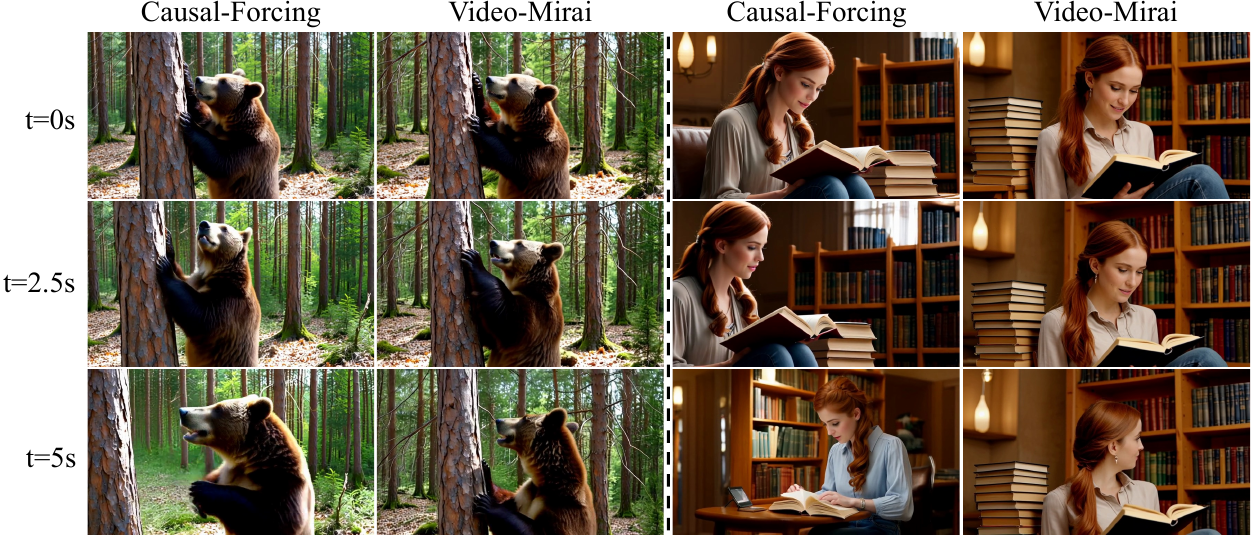}
\caption{\textbf{Qualitative evidence of the planning gap.} Frames at $t{=}0$s, $2.5$s, $5$s from the same prompt. The baseline's local plausibility per segment masks abrupt drift between segments: the bear's motion resets, and the woman's identity and background change. Video-Mirai mitigates these breaks.}
    \label{fig:video_sample}
\end{figure*}

Streaming video generation has a hidden burden: every emitted segment becomes a promise that future segments must keep. Autoregressive (AR) video diffusion follows this interface directly: generate a frame or chunk from the previous context, emit it, and continue. This streaming ability is central to low-latency visual synthesis for interactive world modeling~\cite{bruce2024genie,alonso2024diamond}, game-engine simulation~\cite{valevski2025gamengen}, and embodied intelligence~\cite{chen2024diffusionforcing,teng2025magi,chen2025skyreels}. Yet, it also makes local decisions hard to revise.
A segment may look correct in isolation while failing to specify what must remain true later.
Figure~\ref{fig:video_sample} shows typical failures: in the baseline rollouts, a bear's motion unexpectedly jumps between frames; in another example, the people and the background abruptly change between segments. 
These errors suggest that local visual quality is not sufficient for maintaining identity, layout, and motion over time.

\looseness=-1
We view this as more than generic error accumulation. Output-level supervision is under-constrained: many hidden states can generate nearly the same plausible current segment, but only some retain the information needed for future consistency. The loss tells the model whether its output looks right; it does not directly ask whether the current state contains the information \emph{future} segments will need. We call this missing constraint a representation-level \emph{planning gap}.

This gap points to a simple design principle: use foresight as supervision, not as input. Future evidence is useful for deciding which current states are good, but it cannot be given to a streaming generator at inference. We introduce Video-Mirai, a training-only objective for AR video diffusion that resolves this tension. During training, Video-Mirai lets future segments supervise the current causal state; at inference, the generator remains strictly past-only.
Existing AR video methods improve how models roll out, but they do not close this planning gap. Self-Forcing~\cite{huang2025self} exposes the model to self-generated histories; Causal-Forcing~\cite{zhu2026causal} makes distillation compatible with causal decoding. These methods modify the history distribution and the output target.
Video-Mirai targets a different object: the representation carried by the current causal state.

Video-Mirai is deliberately minimal. The generator first rolls out video causally under the same mask used at inference. A frozen foresight encoder then reads the completed rollout and produces future-informed feature targets for the current states. A lightweight predictor maps each causal hidden state to its corresponding foresight target with a cosine loss. Future segments are used only to construct stopped-gradient supervision; they are never provided as generator inputs. After training, the foresight encoder and predictor are discarded. The deployed generator remains exactly causal, with identical architecture, attention pattern, per-step FLOPs, and KV-cache behavior to the baseline.

Video-Mirai improves both short- and long-horizon generation. On the 5-second VBench, Video-Mirai improves the AR video diffusion baseline Causal-Forcing from $83.82$ to $84.62$ in Total Score. For the 30-second generation, beyond the training horizon, subject consistency improves from $84.93$ to $88.47$, and background consistency from $90.22$ to $91.94$. The method also transfers across AR settings, improving both frame-wise and chunk-wise generation. Beyond generation metrics, representation probes (Figures~\ref{fig:visualization}\ and \ref{fig:quantitative_visualization}) show that future frames become substantially more decodable from frozen Video-Mirai features. Figure~\ref{fig:video_sample} confirms that this internalized foresight translates to visibly more coherent rollouts, mitigating the planning gap in practice.
Unlike prior foresight objectives for image autoregression, Video-Mirai addresses a video-specific planning gap in causal diffusion rollouts, where future supervision must improve temporal coherence while preserving streaming inference and KV-cache-compatible deployment.
In summary, our contributions are threefold:
\begin{itemize}[leftmargin=0.2in]
\setlength{\itemsep}{0em}
\item 
We formulate foresight prediction as a representation-level objective for causal video generation, using future-aware targets during training while preserving strictly causal inference.

\item 
Video-Mirai improves AR video generation across frame-wise and chunk-wise settings, with gains that also extend to 30-second rollouts beyond the training horizon.

\item 
We identify which foresight source, prediction layer, predictor design, and look-ahead window make the training signal effective, and verify through probes that future content becomes more decodable from frozen features.
\end{itemize}

\section{Related Work}

\subsection{Video Diffusion Models}
Large-scale video diffusion models such as Wan~\cite{wan2025wan}, CogVideoX~\cite{yang2024cogvideox}, HunyuanVideo~\cite{kong2024hunyuanvideo}, and MovieGen~\cite{polyak2024moviegen} generate high-fidelity clips by jointly denoising a temporal window, allowing bidirectional interactions among frames within the clip. This full-window interface supports strong short-range consistency, but it is less natural for low-latency settings where frames or chunks should be emitted as soon as they are generated. AR video diffusion instead generates frames or chunks sequentially from past context, enabling streaming inference and supporting rollouts beyond the training horizon, albeit with increasing risk of drift. CausVid~\cite{yin2025causvid}, Diffusion-Forcing~\cite{chen2024diffusionforcing}, MAGI-1~\cite{teng2025magi}, and SkyReels-V2~\cite{chen2025skyreels} exemplify this causal direction. Other approaches extend pretrained video diffusion models to long videos, including StreamingT2V~\cite{henschel2025streamingt2v} and FIFO-Diffusion~\cite{kim2024fifo}. Together, these works expose the tradeoff behind our study: streaming generation requires local causal decisions, while coherent video depends on identity, layout, and motion remaining predictable over future segments.

\subsection{Distillation and Training Techniques for AR Video Diffusion}
To deliver high-quality sampling in a few steps, recent work has focused on distilling bidirectional teachers~\cite{ho2020ddpm,song2021sde,rombach2022ldm,lipman2023flow} into causal students via Distribution Matching Distillation (DMD)~\cite{yin2024dmd,yin2024dmd2}, progressively closing several supervision gaps inherent to this process.
Self-Forcing~\cite{huang2025self} closes the exposure-bias gap by performing AR rollout during training, exposing the student to its own imperfect histories under a holistic video-level loss.
Causal-Forcing~\cite{zhu2026causal} identifies a theoretical flaw in ODE distillation initialization: distilling a bidirectional teacher into a causal student violates frame-level injectivity, thereby preventing the student from faithfully recovering the teacher's flow map. They close this injectivity gap by introducing an AR teacher for ODE initialization.
Rolling-Forcing~\cite{liu2025rolling} targets error accumulation over long horizons through joint denoising across a rolling window, attention-sink anchors, and training over extended non-overlapping windows.
These methods advance the state of the art in causal training stability and past consistency. Our work is orthogonal to theirs. We address the complementary planning gap, namely, a causal generator's inability to anticipate its own future evolution, and our foresight objective can be layered on top of AR video diffusion training paradigms, as our experiments on Self-Forcing and Causal-Forcing demonstrate.

\subsection{Representation Alignment and Foresight}

Aligning a model's intermediate representations with those of a strong external encoder, dating back to knowledge distillation~\cite{hinton2015kd} and intermediate-layer hints~\cite{romero2015fitnets}, has proven more effective for training efficiency and sample quality than pixel-level objectives. REPA~\cite{yu2024repa} regresses mid-layer DiT features onto a pretrained self-supervised encoder such as DINOv2~\cite{oquab2024dinov2}, with follow-ups~\cite{leng2025repae} extending this to end-to-end VAE tuning. A separate line uses representation prediction as a self-supervised pretext task: V-JEPA and its successors~\cite{bardes2024vjepa,assran2025vjepa2}, building on JEPA~\cite{lecun2022path,assran2023ijepa} and masked image modeling~\cite{he2022mae}, predict masked spatiotemporal regions, with the action-conditioned V-JEPA-2-AC training a predictor invoked at inference. The intuition that future-aware signals improve training extends beyond vision, e.g., multi-token prediction~\cite{gloeckle2024mtp} in language models.

Closest in spirit, Mirai~\cite{yu2026mirai} exposes AR \emph{image} generators to future-position information. Video-Mirai extends this to causal \emph{video diffusion}, where the planning gap is sharper: temporal coherence over multi-second rollouts is more demanding than spatial coherence within an image, and the bidirectional/causal architectural asymmetry inherent to video distillation has no counterpart in the image setting.
Video-Mirai draws on REPA-style predictor alignment and JEPA-style latent prediction, recombined for a different purpose: future-aware supervision for a strictly causal video model. Unlike REPA, our target is the temporally shifted hidden state from a bidirectional video model. Unlike V-JEPA, our target is the encoder's representation of the model's own future rollout rather than masked regions, and our predictor is discarded after training. The causal model remains strictly causal at inference. To our knowledge, Video-Mirai is the first to use future-conditioned representation alignment as a training-only signal for causal video diffusion.

\section{Method}
\label{sec:method}

\subsection{Preliminaries}
\label{sec:method:prelim}

We consider AR video diffusion that distills a pretrained bidirectional diffusion model into a few-step causal generator $G_\theta$. A video is a sequence of temporal segments $\mathbf{x} = \{\mathbf{X}_1, \dots, \mathbf{X}_N\}$, where a segment is a single latent frame (frame-wise) or a chunk of consecutive latent frames (chunk-wise). The causal constraint requires:
\begin{equation}
p_\theta(\mathbf{x}) \;=\; \prod_{i=1}^{N} p_\theta(\mathbf{X}_i \mid \mathbf{X}_{<i}),
\label{eq:causal_factorization}
\end{equation}

so segment $\mathbf{X}_i$ is generated from $\mathbf{X}_{<i}$ alone, with no access to $\mathbf{X}_{>i}$ at inference. Standard distillation pipelines such as Causal-Forcing~\cite{zhu2026causal} realize this through three stages: AR teacher fine-tuning, causal ODE distillation, and asymmetric DMD~\cite{yin2024dmd,yin2024dmd2}. We adopt this pipeline as our base.

The factorization in Eq.~\ref{eq:causal_factorization} only constrains each $\mathbf{X}_i$ to be plausible given $\mathbf{X}_{<i}$. It says nothing about whether the hidden state $\mathbf{h}_i^L$ used to generate $\mathbf{X}_i$ retains the information needed to generate \emph{future} segments coherently. For example, a per-segment training objective can be written as follows:

\begin{equation}
\mathcal{L}_{\text{present}}(\theta) \;=\; \mathbb{E}_{\mathbf{x}} \sum_{i=1}^{N} 
\ell\!\left( G_\theta(\mathbf{X}_i \mid \mathbf{X}_{<i}) \right),
\label{eq:present_loss}
\end{equation}

where $\ell$ is any per-segment supervision applied to the model's output distribution (DMD score, flow-matching, etc.). Existing objectives supervise generated outputs, including entire video rollouts, but do not explicitly align intermediate causal states with future-informed representation targets. Video-Mirai adds this representation-level supervision.
Many distinct hidden states $\mathbf{h}_i^L$ can drive nearly the same $\mathbf{X}_i$ to optimum~\cite{george2021clone,swaminathan2023schema}: some retain identity, layout, and motion cues that future segments will need; others discard them. The loss has no preference between the two~\cite{dedieu2024cognitive}. We call this missing constraint the \emph{representation-level planning gap}: output-level losses do not ask whether $\mathbf{h}_i^L$ is a good state from which to continue. 
Recent advances in AR video diffusion address adjacent yet distinct gaps, including exposure bias~\cite{huang2025self}, ODE-distillation injectivity~\cite{zhu2026causal}, and long-horizon error accumulation~\cite{liu2025rolling}. None of them, however, explicitly supervises $\mathbf{h}_i^L$. 
They reshape the history distribution, the distillation initialization, or the rollout protocol, not the supervision applied to the causal state itself. The planning gap is therefore orthogonal to all three. Closing it requires supervising $\mathbf{h}_i^L$ with a \emph{foresight} signal: a target derived from the model's own future rollout.

\subsection{Video-Mirai}
\label{sec:method:video-mirai}

\begin{figure*}[t]
    \centering
    \includegraphics[width=1\linewidth]{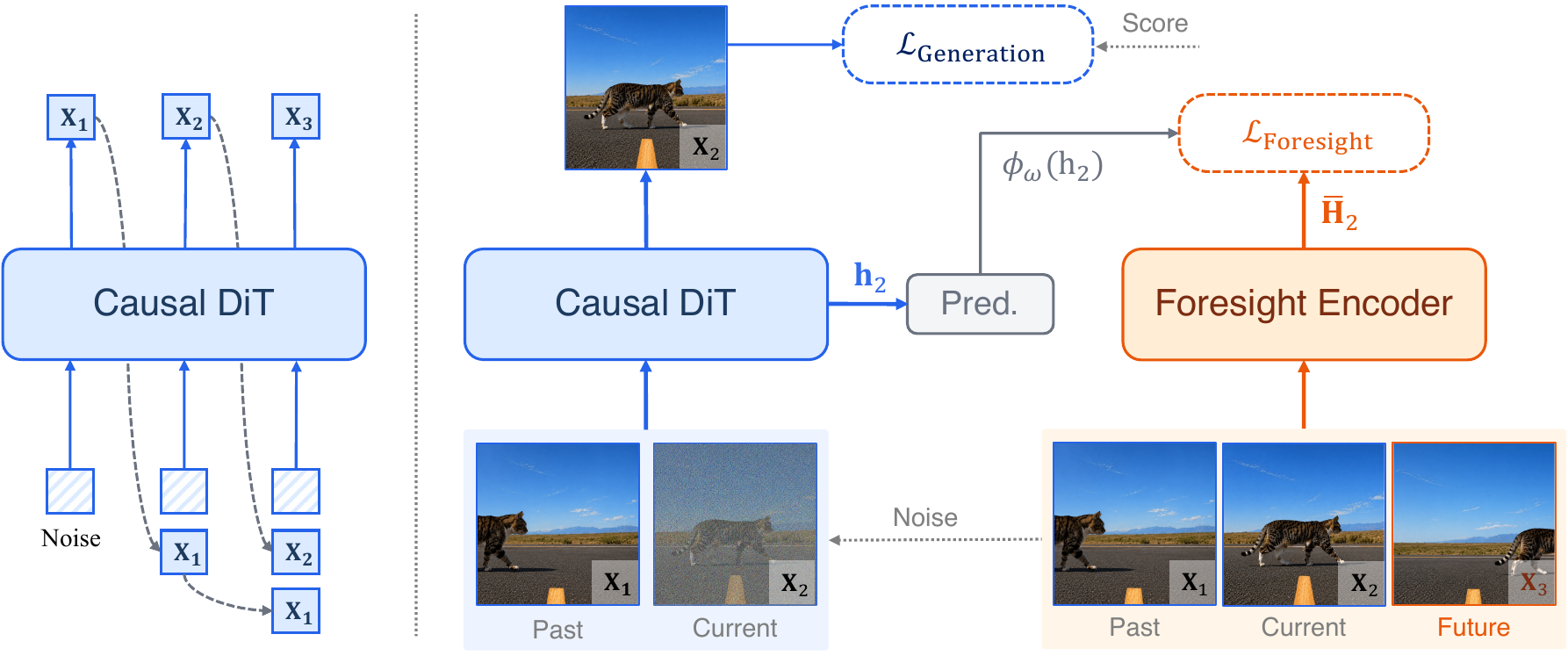}
    \caption{\textbf{Overview of Video-Mirai.} We take a three-segment video as an example. The causal DiT denoises $\mathbf{X}_2$ from its noisy version 
conditioned on $\mathbf{X}_1$ via KV-cache. A frozen foresight encoder processes the causal DiT's clean rollout $\{\mathbf{X}_1, \mathbf{X}_2, \mathbf{X}_3\}$, including the future segment ${\mathbf{X}}_3$. A predictor maps the causal DiT's hidden state $\mathbf{h}_2$ into the encoder's space, where the foresight loss aligns it with the encoder's fused hidden state $\bar{\mathbf{H}}_2$, which contains foresight information.}
    \label{fig:method}
\end{figure*}
Figure~\ref{fig:method} illustrates an overview of Video-Mirai: while denoising the current segment, the causal DiT generator's mid-depth feature is fed to a predictor $\phi_\omega$ to predict the foresight encoder's feature on a future segment of the same rollout, via a cosine similarity loss:
\begin{equation}
\ell^{\text{F}}_i(\delta) \;=\; 1 - \cos\!\big(\phi_\omega(\mathbf{h}_i^{L}),\; \mathrm{sg}[\mathbf{H}_{i+\delta}^{L'}]\big).
\label{eq:foresight_single}
\end{equation}

Let $\mathbf{h}_i^{L} \in \mathbb{R}^{T \times d_c}$ denote the causal generator's hidden state at layer $L$ when generating $\mathbf{X}_i$, and $\mathbf{H}_{i+\delta}^{L'} \in \mathbb{R}^{T \times d_f}$ the foresight encoder's hidden state on the causal generator's rollout segment $\mathbf{X}_{i+\delta}$ at a matched mid-depth layer $L'$, computed when the encoder processes the full rollout $\mathbf{x}$. Two design questions follow immediately: (i) is foresight useful, and how far ahead should the alignment look, and (ii) should multiple offsets be combined by fusing targets or by averaging losses. We answer both before assembling the full objective.

\textbf{Foresight window.}  
We compare four offset configurations: current only ($\{0\}$), one-segment ahead only ($\{1\}$), current plus one-segment ($\{0,1\}$), and current plus two-segment ($\{0,1,2\}$). In the chunk-wise setting, $\delta{=}1$ corresponds to a 3-latent-frame look-ahead. As shown in Table~\ref{tab:foresight_window}, current plus one-segment ($\{0,1\}$) yields the best Quality and Total Scores. Crucially, even current-only alignment carries foresight information because the foresight encoder here is bidirectional, so its hidden state on the current chunk is already conditioned on future chunks. Thus, foresight arises not only from explicit future-offset alignment but also from implicit future information through bidirectional attention. Extending the window to $\{0,1,2\}$ further improves Semantic but degrades Quality, indicating that longer-range alignment trades visual fidelity for global coherence.

\textbf{Target fusion vs.\ per-offset losses.}  
Given a window $\Delta = \{0, 1, \dots, K\}$ ($K{=}1$ by default), we can either compute one loss per offset and average, or fuse the encoder's features across $\Delta$ into a single target and align once. Table~\ref{tab:projector} compares four predictor designs spanning these two regimes: the pointwise MLP and the DiT predictor, each on either the fused target or a per-offset variant.
Two findings emerge. First, fusing across offsets combines features from the current and future segments of the same rollout into a single target, and Table~\ref{tab:projector} shows that this approach outperforms averaging per-offset losses: the single-head MLP beats Multi-MLP, and the single-head DiT beats DiT-AdaLN. Second, DiT outperforms the pointwise MLP at matched depth, suggesting the alignment benefits from token mixing. We also vary DiT depth among $\{2, 3, 4\}$ blocks and find that 3 blocks give the best Total Score.
We therefore adopt a 3-block DiT with target fusion as our default. With this choice, the fused target and per-segment loss are:
\begin{equation}
\bar{\mathbf{H}}_i \;=\; \frac{1}{|\Delta|} \sum_{\delta \in \Delta} \mathbf{H}_{i+\delta}^{L'},
\quad
\ell^{\text{F}}_i \;=\; 1 - \cos\!\big(\phi_\omega(\mathbf{h}_i^{L}),\; \mathrm{sg}[\bar{\mathbf{H}}_i]\big).
\label{eq:foresight}
\end{equation}

\begin{table}[t]
\centering
\caption{\textbf{Foresight window comparison.} Effect of the offset set $\Delta$ on top of Causal-Forcing. }
\label{tab:foresight_window}
\begin{tabular}{lcccc}
\toprule
Setting & $\Delta$ & Quality $\uparrow$ & Semantic $\uparrow$ & Total $\uparrow$ \\
\midrule
Causal-Forcing & --        & 84.54  & 80.93  & 83.82 \\
\midrule
Current only            & $\{0\}$         & 85.07  & 81.55  & 84.36 \\
1-segment ahead only               & $\{1\}$         & 84.82  & 81.50  & 84.15 \\
Current + 1-segment ahead  & $\{0, 1\}$      & \textbf{85.38}  & 81.59  & \textbf{84.62} \\
Current + 2-segment ahead  & $\{0, 1, 2\}$   & 85.11  & \textbf{82.00}  & 84.49 \\
\bottomrule
\end{tabular}
\end{table}

\begin{table}[t]
  \centering
  \caption{\textbf{Predictor architecture comparison.}
  Fusion means encoder targets are averaged across $\Delta$ before the loss; per-offset means each offset is handled separately.}
  \label{tab:projector}
  \setlength{\tabcolsep}{6pt}
  \begin{tabular}{lcccccc}
    \toprule
    Architecture     & Target handling & Depth & Quality $\uparrow$ & Semantic $\uparrow$ & Total $\uparrow$ \\
    \midrule
    Causal-Forcing  & --        & --    & 84.54  & 80.93           & 83.82 \\
    \midrule
    MLP              & Fusion           & 3     & 85.09  & 81.58           & 84.38 \\
    Multi-MLP        & Per-offset       & 3     & 85.00  & 81.12           & 84.22 \\
    \midrule
    DiT & Fusion & 2 & 85.14 & \textbf{81.85} & 84.48 \\
    DiT              & Fusion           & 3     & 85.38  & 81.59           & \textbf{84.62} \\
    DiT              & Fusion           & 4     & \textbf{85.41} & 81.19   & 84.57 \\
    DiT-AdaLN        & Per-offset       & 3     & 85.34  & 81.46           & 84.56 \\
    \bottomrule
  \end{tabular}
\end{table}
\textbf{Training.}  
At each step, the causal generator unrolls a video $\mathbf{x}$ segment by segment from noise, following the standard few-step denoising loop of Causal-Forcing with the usual KV-cache, and we cache its mid-depth states $\{\mathbf{h}_i^{L}\}$. The frozen foresight encoder then processes the full rollout $\mathbf{x}$ in a single forward pass at zero diffusion timestep, yielding $\{\mathbf{H}_{i}^{L'}\}$. 
For every segment with a near-future target available ($i \le N - K$), $\phi_\omega(\mathbf{h}_i^{L})$ is pulled toward the fused target $\bar{\mathbf{H}}_i$ via Eq.~\ref{eq:foresight}, combined with the asymmetric DMD generation loss:
\vspace{-2mm}
\begin{equation}
\mathcal{L}_{\text{total}} = \mathcal{L}_{\text{Generation}} + \lambda \, 
\underbrace{\frac{1}{N - K} \sum_{i=1}^{N - K} \ell^{\text{F}}_i}_{\mathcal{L}_{\text{Foresight}}},
\label{eq:total_loss}
\end{equation}

\vspace{-3mm}

where we use $\lambda = 0.2$ by default.
The DMD term scores a re-noised version of the same rollout against a frozen Wan-14B real-score teacher and a Wan-1.3B fake-score critic, following Self-Forcing~\cite{huang2025self}. The remaining designs, such as the foresight encoder, injection depth $L$, and loss form, are ablated in \S\ref {sec:exp:component_wise}. We also discuss applying Video-Mirai in a different stage in Appendix~\ref{sec:appendix:stages}.

\section{Experimental Results}
\label{sec:exp}

\subsection{Setup}

\paragraph{Implementation details.} We use Wan2.1-T2V-1.3B~\cite{wan2025wan} (henceforth Wan-1.3B) as the causal generator backbone, which generates 81-frame videos at a resolution of 832 × 480. Our main results are reported on top of the three-stage Causal-Forcing~\cite{zhu2026causal} pipeline under the chunk-wise setting, where each segment is a chunk of 3 consecutive latent frames. The training procedure in every stage follows Causal-Forcing. We also use Video-Mirai as a drop-in addition to Self-Forcing~\cite{huang2025self} and to the frame-wise Causal-Forcing variant. 
Unless otherwise noted, Video-Mirai uses Wan2.1-T2V-14B (henceforth Wan-14B) as the foresight encoder. All distillation runs use $8\times$ H100 GPUs with gradient accumulation 8. The training prompts come from the filtered VidProM \cite{wang2024vidprom} extension released by Self-Forcing. Other training settings follow the baseline's recipe. More details are in Appendix \ref{sec:appendix:impl}.

\paragraph{Evaluation.}
We evaluate on VBench~\cite{huang2024vbench}. For 5-second generation, each score is the mean over 5 generated videos per prompt across the full VBench prompt set, following the benchmark's standard protocol. For the 30-second long-horizon evaluation, we roll out the same trained checkpoints on 200 randomly selected MovieGen \cite{polyak2024moviegen} prompts, following the Rolling-Forcing~\cite{liu2025rolling} protocol, and report the subset of VBench dimensions that are computable under its prompt-agnostic protocol.
Due to the high cost of video distillation, we follow prior work and report paired prompt-level bootstrap significance instead of variance over multiple training seeds. Details are provided in Appendix \ref{sec:appendix:stats}.

\subsection{Component-wise Analysis}

\label{sec:exp:component_wise}

\begin{table}[t]
\centering
\caption{\textbf{Injection depth comparison.} Each row pairs a causal generator layer with a foresight encoder layer at matched relative depth $\alpha$.}
\label{tab:depth}
\setlength{\tabcolsep}{8pt}
\begin{tabular}{lccc}
  \toprule
  Layers ($\alpha$) & Quality $\uparrow$ & Semantic $\uparrow$ & Total $\uparrow$ \\
  \midrule
  Causal-Forcing & 84.54 & 80.93 & 83.82 \\
  \midrule
  9$\to$12 (0.3)  & 85.14 & 81.37 & 84.39 \\
  15$\to$20 (0.5) & \textbf{85.38} & \textbf{81.59} & \textbf{84.62} \\
  24$\to$32 (0.8) & 84.77 & 81.54 & 84.12 \\
  \bottomrule
\end{tabular}
\end{table}

\begin{table}[t]
\centering
\caption{\textbf{Foresight encoder comparison.} Total Score on 5-second VBench. \emph{Implicit}: bidirectional attention lets the target see later segments; \emph{explicit}: the target includes the next segment. All
variants use the default 3-block DiT predictor, layered on Causal-Forcing chunk-wise.}
\label{tab:foresight_encoder}
\setlength{\tabcolsep}{8pt}
\renewcommand{\arraystretch}{1.15}
\small
\begin{tabular}{llllc}
  \toprule
  Foresight encoder & Temporal attention & Future access & Offsets $\Delta$ & Total $\uparrow$ \\
  \midrule
  Causal-Forcing & -- & -- & -- & 83.82 \\
  \midrule
  \multirow{2}{*}{EMA}      & \multirow{2}{*}{causal}        & none                & $\{0\}$   & 83.71 \\
                            &                                & explicit            & $\{0,1\}$ & 83.88 \\
  \midrule
  \multirow{2}{*}{Wan-1.3B} & \multirow{2}{*}{bidirectional} & implicit            & $\{0\}$   & 83.86 \\
                            &                                & implicit + explicit & $\{0,1\}$ & 83.96 \\
  \midrule
  \multirow{2}{*}{Wan-14B}  & \multirow{2}{*}{causal}        & none                & $\{0\}$   & 84.06 \\
                            &                                & explicit            & $\{0,1\}$ & 84.53 \\
  \midrule
  \multirow{2}{*}{Wan-14B}  & \multirow{2}{*}{bidirectional} & implicit            & $\{0\}$   & 84.36 \\
                            &                                & implicit + explicit & $\{0,1\}$ & \textbf{84.62} \\
  \bottomrule
\end{tabular}
\end{table}

\paragraph{Injection depth.} Foresight injection requires choosing both a causal generator layer $L$ and an encoder layer $L'$  . Since the causal generator (30 blocks) and the encoder (40 blocks) differ in depth, we pair layers at matched relative depth: $L = 30\alpha$, $L' = 40\alpha$. 
Table~\ref{tab:depth} shows that mid-depth ($\alpha{=}0.5$) is best. Shallow layers have not yet accumulated the abstract planning features that benefit from foresight, while deep layers are close to pixel-level reconstruction and have less freedom to be reshaped by an auxiliary objective. We adopt $\alpha{=}0.5$ throughout.

\paragraph{Foresight encoder.}  To examine how much the future contributes, we compare foresight encoders of different sizes with and without access to it: the EMA copy of the causal generator itself, the bidirectional Wan-1.3B and Wan-14B models, and Wan-14B under a 3-latent-frame chunk-wise causal mask that hides later segments.
The results in Table~\ref{tab:foresight_encoder} show that without future access, alignment does not reproduce the gain: the causal EMA falls below the baseline (83.71 vs.\ 83.82), and the causal Wan-14B reaches only 84.06. Explicit future access improves every encoder, implicit access lifts Wan-14B from 84.06 to 84.36, and combining both is best (84.62). Encoder capacity matters as well: the causal Wan-14B (84.06) exceeds the bidirectional Wan-1.3B with explicit future access (83.96).
We adopt Wan-14B as our default, since it is the strongest encoder and is already the real-score model in the baseline DMD pipeline. A much smaller image encoder from outside the model family, DINO ViT-S/16~\cite{caron2021dino}, also improves the baseline (Appendix~\ref{sec:appendix:dino}).

\begin{figure*}[t]
  \centering
  \begin{minipage}[b]{0.49\linewidth}
    \centering
    \includegraphics[height=4.8cm]{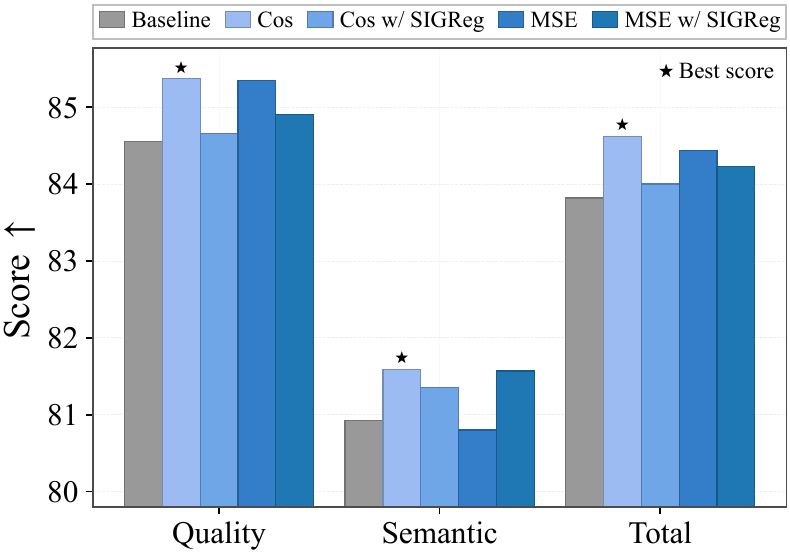}
  \end{minipage}%
  \hfill
  \begin{minipage}[b]{0.49\linewidth}
    \centering
    \includegraphics[height=4.8cm]{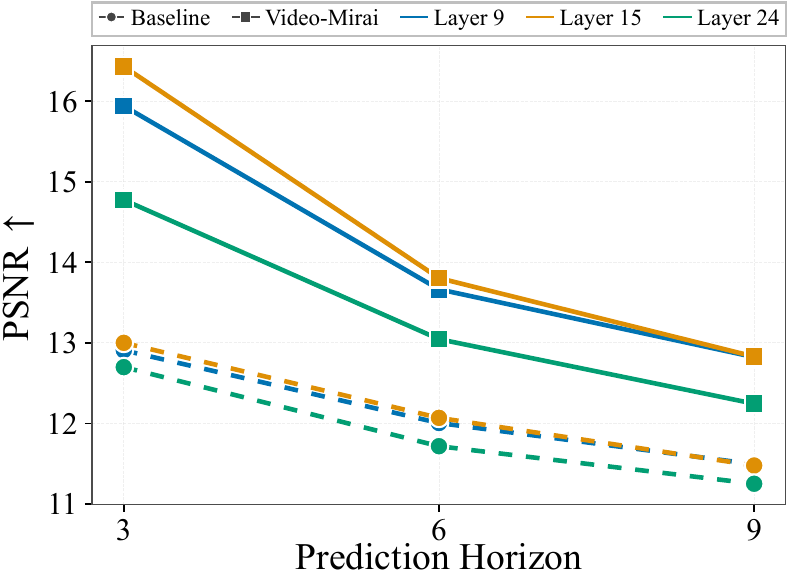}
  \end{minipage}
  \par\vspace{-3mm}
  \begin{minipage}[t]{0.49\linewidth}
    \caption{\textbf{Foresight loss comparison.} Cosine 
    similarity vs.\ MSE, with and without the SIGReg.}
    \label{fig:loss_comparison}
  \end{minipage}%
  \hfill
  \begin{minipage}[t]{0.49\linewidth}
    \caption{\textbf{Future-frame readout fidelity across layers and horizons.} }
    \label{fig:quantitative_visualization}
  \end{minipage}
  \vspace{-2.mm}
\end{figure*}
\paragraph{Foresight loss.}
Figure~\ref{fig:loss_comparison} compares different foresight losses with or without a Sketched Isotropic Gaussian Regularization term (SIGReg)~\cite{balestriero2025lejepa}. 
Cosine similarity alone works best. We attribute this to two effects. First, MSE forces the generator's feature to match the encoder's
direction, scale, and full coordinate distribution, pulling it away from its native feature geometry and dropping the Semantic Score. 
Second, adding SIGReg consistently degrades the Total Score under both losses, indicating that explicitly shaping the projected feature distribution toward isotropic Gaussianity conflicts with the alignment objective. We adopt cosine similarity without SIGReg as the default. Loss weight ablations are in Appendix~\ref{sec:appendix:loss_weight}.

\subsection{Foresight Visualization}
\label{sec:exp:probing}

All evidence so far is behavioral: VBench scores improve after foresight training. A sharper question is whether foresight is
actually \emph{internalized} into the causal generator's representations. We design a representational probing experiment to answer this directly.

\paragraph{Probe setup.}  
We take two frozen causal generators with identical architectures: the Causal-Forcing baseline and Video-Mirai. For each model, we train a small MLP decoder to reconstruct the clean RGB future frame $\delta$ chunks ahead from the model's layer-15 hidden state on the \emph{current} chunk. No foresight predictor is used at probing time; the decoder is purely a readout. Decoders for the two models are trained separately within an identical budget. If Video-Mirai has internalized future information, its features should support substantially more faithful future reconstructions.

\paragraph{Qualitative and quantitative readout.}  
Figure~\ref{fig:visualization} shows three probe samples.
We can see that Video-Mirai faithfully reconstructs the future frame, whereas the baseline readout remains closer to the current frame, as expected for a feature that lacks foresight.
We then quantify the readout fidelity at scale across three feature-extraction layers $L\in\{9, 15, 24\}$ and three prediction horizons of $\{3, 6, 9\}$ latent frames ahead, which means $\delta\in\{1, 2, 3\}$ chunks, reporting MSE, PSNR, and LPIPS. For each layer, we train a separate MLP decoder on the causal generator's hidden state and evaluate on held-out prompts. 
For brevity, we plot PSNR in Figure~\ref{fig:quantitative_visualization}. The MSE and LPIPS curves, which show the same trend, are provided in Appendix~\ref{sec:appendix:probing}. Video-Mirai consistently outperforms the baseline at every layer and every horizon across all three metrics. 
The gap is largest at near horizons, where foresight supervision applies most directly, and persists at a horizon of 9 latent frames ($\delta{=}3$ chunks), well beyond the training-time foresight window, indicating that foresight internalization generalizes rather than memorizes.

\begin{figure*}[t]
  \centering
    \includegraphics[width=\linewidth]{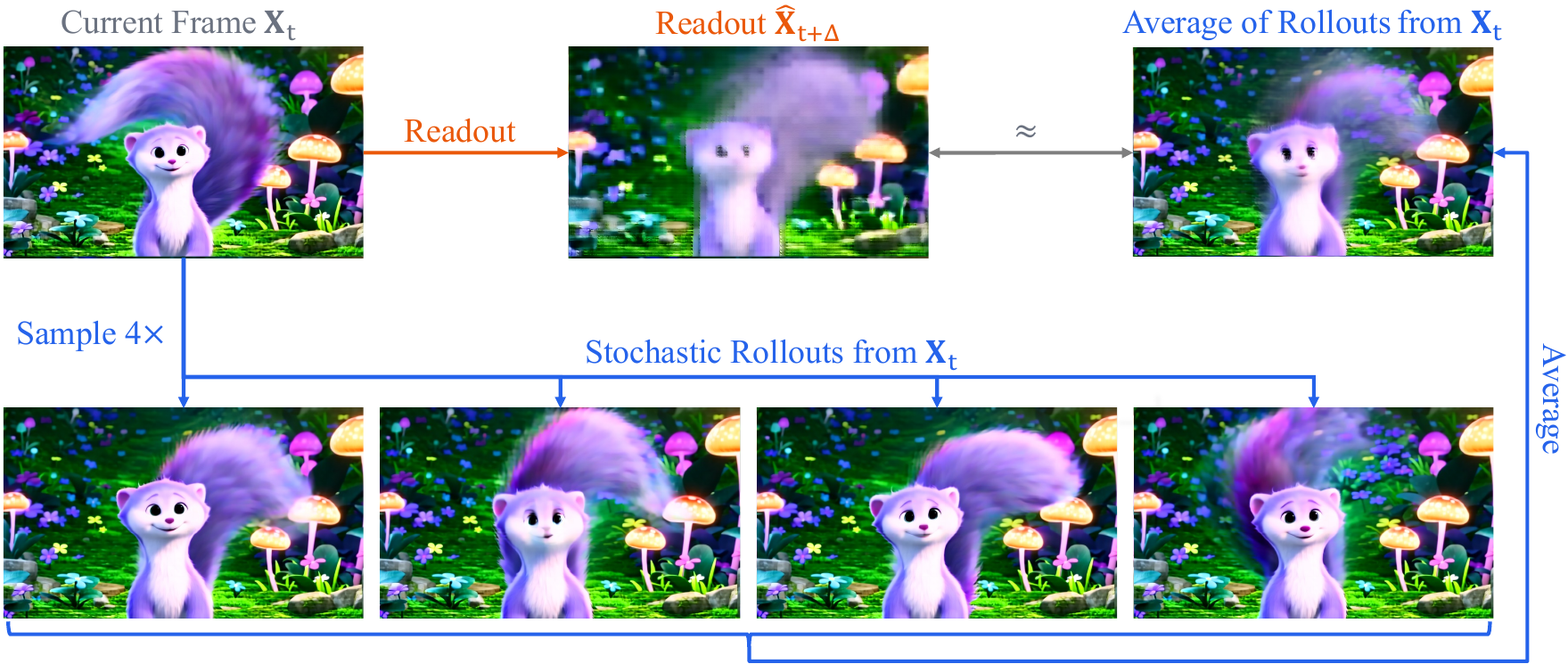}
      \vspace{-5mm}
\caption{\textbf{The readout resembles the average of multiple future rollouts.}}
   \label{fig:visualization_4_future}
\end{figure*}

\paragraph{Distribution of futures.}  
The probes above show that Video-Mirai's features support future readout, but the future of a given current is not deterministic. Running the causal generator from the same current with different noise seeds yields a distribution of valid futures. So do the features encode \emph{a} specific future, or the \emph{distribution} of futures? 
Figure~\ref{fig:visualization_4_future} visualizes this. We sample four independent rollouts from the same current under different noise seeds, confirming that the future is genuinely stochastic. For example, the purple ermine's tail ends up at different positions across rollouts.
The readout from Video-Mirai's current segment resembles the \emph{average} of these four rollouts. The readout can already capture identity, layout, and motion statistics of the future before any of those future frames are actually generated.
This suggests that Video-Mirai's mid-depth features summarize what is about to happen rather than a single sampled continuation.

\subsection{System-Level Comparison}
\label{sec:exp:system}

\begin{table}[!t]
{\renewenvironment{table}[1][]{}{\par}%

\begin{table}[!t]
\centering
\small
\caption{\textbf{Quantitative comparisons with existing methods.} On 5-second VBench, Video-Mirai achieves the best Quality, Semantic, and Total Scores while preserving the throughput and latency of the underlying AR baseline. Bold marks the best throughput and latency, and Video-Mirai scores above their baseline.}
\label{tab:comparison}
\setlength{\tabcolsep}{2.8pt}
\renewcommand{\arraystretch}{1.1}
\begin{tabular}{lccccccc}
\toprule
Model & \#Params & Resolution & \makecell{Throughput\\(FPS) $\uparrow$} & \makecell{Latency\\(s) $\downarrow$} & \makecell{Quality\\Score $\uparrow$} & \makecell{Semantic\\Score $\uparrow$} & \makecell{Total\\Score $\uparrow$} \\
\midrule
\rowcolor{gray!20}
\multicolumn{8}{l}{\textit{Bidirectional Video Diffusion Models}} \\
LTX-Video~\cite{hacohen2024ltx} & 1.9B & 768$\times$512 & 8.98 & 13.5 & 81.88 & 71.62 & 79.83 \\
Wan2.1~\cite{wan2025wan}        & 1.3B & 832$\times$480 & 0.78 & 103  & 84.30 & 79.65 & 83.37 \\
\midrule
\rowcolor{gray!20}
\multicolumn{8}{l}{\textit{Autoregressive Video Diffusion Models}} \\
NOVA~\cite{deng2024nova}             & 0.6B & 768$\times$480 & 0.88 & 4.1 & 80.66 & 78.92 & 80.31 \\
Pyramid Flow~\cite{jin2024pyramidal} & 2B   & 640$\times$384 & 6.70 & 2.5 & 83.41 & 70.11 & 80.75 \\
SkyReels-V2~\cite{chen2025skyreels}  & 1.3B & 960$\times$540 & 0.49 & 112 & 83.96 & 74.01 & 81.97 \\
MAGI-1~\cite{teng2025magi}           & 4.5B & 832$\times$480 & 0.19 & 282 & 81.67 & 67.72 & 78.88 \\
\midrule
\rowcolor{gray!20}
\multicolumn{8}{l}{\textit{Distilled Autoregressive Video Models}} \\
CausVid~\cite{yin2025causvid}     & 1.3B & 832$\times$480 & \textbf{17.0} & 0.69 & 83.98          & 70.72          & 81.33 \\
\midrule
Self-Forcing~\cite{huang2025self} (chunk-wise)& 1.3B & 832$\times$480 & \textbf{17.0} & 0.69 & 84.37 & 80.87 & 83.67 \\
 \rowcolor{gray!10}
    \;\textbf{+ Video-Mirai (Ours)}   & 1.3B & 832$\times$480 & \textbf{17.0} & 0.69 & \textbf{84.82} & \textbf{81.45} & \textbf{84.15}  \\
\midrule
Causal-Forcing~\cite{zhu2026causal} (frame-wise) & 1.3B & 832$\times$480 &  8.9 & \textbf{0.45}  & 83.16 & 78.73 & 82.27 \\
 \rowcolor{gray!10}
    \;\textbf{+ Video-Mirai (Ours)}   & 1.3B & 832$\times$480 &  8.9 & \textbf{0.45}  &  \textbf{84.59} & \textbf{79.66} & \textbf{83.60} \\
\midrule
Causal-Forcing~\cite{zhu2026causal} (chunk-wise) & 1.3B & 832$\times$480 & \textbf{17.0} & 0.69 & 84.54 & 80.93 & 83.82 \\
 \rowcolor{gray!10}
    \;\textbf{+ Video-Mirai (Ours)}       & 1.3B & 832$\times$480 & \textbf{17.0} & 0.69 & \textbf{85.38} & \textbf{81.59} & \textbf{84.62} \\
\bottomrule
\end{tabular}
\end{table}%
\par\vspace{14pt}%
\begin{table}[t]
\centering
\small
\caption{\textbf{Transfer to another AR model.} SkyReels-V2-DF-1.3B fine-tuned for 500 steps with and without the foresight loss.}
\label{tab:skyreels}
\setlength{\tabcolsep}{8pt}
\begin{tabular}{lccc}
\toprule
Method & Quality $\uparrow$ & Semantic $\uparrow$ & Total $\uparrow$ \\
\midrule
SkyReels-V2-DF-1.3B (fine-tuned) & 83.09 & 76.72 & 81.82 \\
\rowcolor{gray!10}
\;\textbf{+ Video-Mirai (Ours)} & \textbf{84.26} & \textbf{77.42} & \textbf{82.89} \\
\bottomrule
\end{tabular}
\end{table}
}%
\vspace{-2mm}%
\end{table}
Table~\ref{tab:comparison} situates Video-Mirai within the broader landscape of video diffusion models, comparing against bidirectional, AR, and distilled-AR video diffusion models on VBench. 
We apply Video-Mirai as a drop-in addition on top of three distilled-AR baselines: Self-Forcing chunk-wise and Causal-Forcing in both frame-wise and chunk-wise variants. The results show that Video-Mirai improves Quality, Semantic, and Total Scores at every baseline, confirming that foresight internalization transfers across distillation frameworks and across frame-wise and chunk-wise generation. Figures~\ref{fig:video_sample} and \ref{fig:video_sample_rest} show representative qualitative comparisons. Additional examples are in Appendix \ref{sec:appendix:comparison_results}.

Beyond these distilled students, we also apply the foresight loss to SkyReels-V2-DF-1.3B~\cite{chen2025skyreels}, a released AR video model with different pretraining data and training objective. We fine-tune the official checkpoint for 500 teacher-forcing steps at batch size 32 with and without the foresight loss, keeping data and seeds identical, and evaluate on all 946 VBench prompts with one video per prompt. As shown in Table~\ref{tab:skyreels}, the foresight loss improves Quality, Semantic, and Total Scores, raising Total from 81.82 to 82.89.

\begin{figure*}[t]
  \centering
    \includegraphics[width=\linewidth]{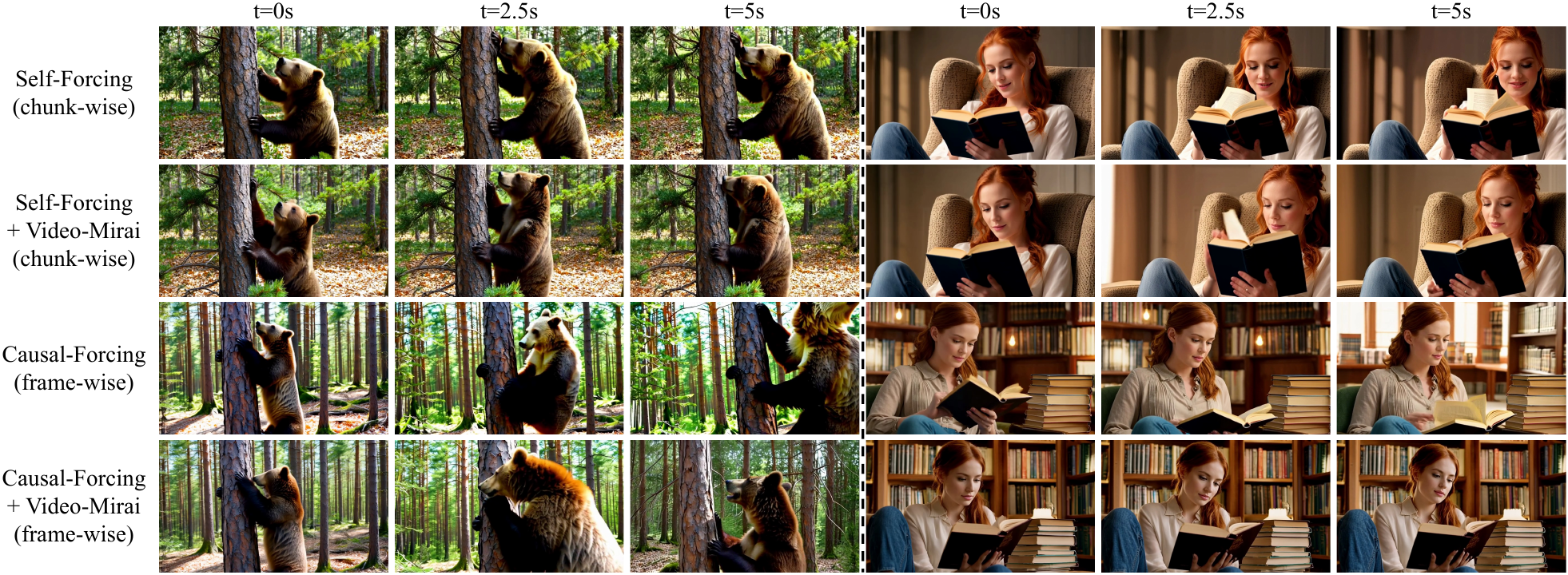}
    \caption{\textbf{Qualitative comparison.} Representative frames at $t{=}0$s, $2.5$s, $5$s from the rollouts of the same prompt under the baselines and their Video-Mirai counterparts. Video-Mirai rollouts preserve subject identity and scene composition more reliably across time.}
    \label{fig:video_sample_rest}
    \vspace{-4mm}
\end{figure*}
\begin{table*}[!t]
  \centering
  \small
  \caption{\textbf{System-level comparison on VBench.} Left: 5-second generation. Right: 30-second long-horizon generation. Video-Mirai is applied as a drop-in addition on top of chunk-wise Self-Forcing and both frame-wise and chunk-wise Causal-Forcing.}
  \vspace{1mm}
  \label{tab:system_comparison_consistency}
  \setlength{\tabcolsep}{3pt}
  \begin{tabular}{l ccc ccc}
    \toprule
& \multicolumn{3}{c}{5-second} & \multicolumn{3}{c}{30-second} \\
\cmidrule(lr){2-4} \cmidrule(lr){5-7}
Method & \makecell{\scriptsize Subject\\\scriptsize Consistency} $\uparrow$ & \makecell{\scriptsize Background\\\scriptsize Consistency} $\uparrow$ & \makecell{\scriptsize Overall\\\scriptsize Consistency} $\uparrow$ 
       & \makecell{\scriptsize Subject\\\scriptsize Consistency} $\uparrow$ & \makecell{\scriptsize Background\\\scriptsize Consistency} $\uparrow$ & \makecell{\scriptsize Overall\\\scriptsize Consistency} $\uparrow$ \\
    \midrule
    Self-Forcing (chunk-wise)        & 96.18 & 96.43 & 26.70 & 89.83 & 92.72 & 25.02 \\
    \rowcolor{gray!10}
    \;\textbf{+ Video-Mirai (Ours)}       &  \textbf{96.77} & \textbf{96.85} & \textbf{26.84} & \textbf{91.62} & \textbf{93.77} & \textbf{25.33} \\
        \midrule
    Causal-Forcing (frame-wise)       & 90.67 & 92.97 & 26.42 &  75.60 & 84.41 & 23.25 \\
    \rowcolor{gray!10}
    \;\textbf{+ Video-Mirai (Ours)}      & \textbf{93.13} & \textbf{94.12} & \textbf{26.57} & \textbf{76.90} &  \textbf{85.07} & \textbf{23.66} \\
    \midrule
    Causal-Forcing (chunk-wise)        & 96.05 & 95.92 & 26.83  & 84.93 & 90.22 & 24.93 \\
    \rowcolor{gray!10}
    \;\textbf{+ Video-Mirai (Ours)}       & \textbf{96.41} & \textbf{96.54} & \textbf{26.85}  & \textbf{88.47} & \textbf{91.94} & \textbf{25.03} \\
    \bottomrule
  \end{tabular}
\vspace{-2mm}
\end{table*}

The improvement extends to 30-second long-horizon generation, on precisely the dimensions tied to causal short-sightedness:
subject and background consistency improve substantially over the no-foresight baseline, and overall consistency improves slightly, as shown in Table~\ref{tab:system_comparison_consistency}. In chunk-wise settings, Video-Mirai yields larger gains on all three consistency metrics in the 30-second evaluation than in the 5-second evaluation. These results support its effectiveness beyond the training horizon, although the two evaluations use different prompt sets. The gains grow further at 60 seconds (Appendix~\ref{sec:appendix:movie60s}).
This performance is achieved at identical inference cost, since the foresight encoder and predictor are discarded after training. 
Training-time overhead is analyzed in Appendix~\ref{sec:appendix:flops}.

\section{Conclusion}

We introduced Video-Mirai, a training-only paradigm that closes the representation-level planning gap in causal AR video diffusion: a single cosine alignment loss pulls the causal generator's mid-depth features toward a frozen bidirectional encoder's view of its own rollout. At inference, the encoder and predictor are discarded, leaving the architecture, per-step FLOPs, and KV-cache behavior unchanged. 
As a drop-in addition to AR video models, Video-Mirai lifts Causal-Forcing's VBench scores at 5 seconds and widens the margin at 30-second generation. Representation probes confirm that future information is internalized into the causal weights, turning anticipation into a property of the causal forward pass. Limitations are discussed in Appendix \ref{sec:limitations}. Our results argue that causality is a constraint on inference, not on what training can teach a model to anticipate.

\newpage

\newpage   %
\section*{Acknowledgments}
This research is partially supported by JSPS KAKENHI Grant Number JP26K21317 and the Initiative on Recommendation Program for Young Researchers and Woman Researchers, Information Technology Center, The University of Tokyo.
The computations were partly performed on the Miyabi-G supercomputer at the Information Technology Center, The University of Tokyo.

\bibliographystyle{plain}
\bibliography{reference}
\newpage
\appendix
\pagebreak

\clearpage
\begin{center}
    \Huge \textbf{Appendix}
\end{center}
\section{Implementation Details}
\label{sec:appendix:impl}

\paragraph{Model architecture.}
The causal generator backbone is Wan2.1-T2V-1.3B~\cite{wan2025wan}: 30 transformer
blocks, hidden dimension 1536, FFN dim 8960, 12 heads, with QK-normalization
and causal attention along the temporal axis. The foresight encoder
is the frozen Wan2.1-T2V-14B: 40 blocks, hidden dimension 5120, FFN dim 13,824, 40 heads, with bidirectional attention. We extract the causal generator's hidden state at layer 15 ($\alpha = 0.5$) and the encoder's hidden state at layer
20 ($\alpha = 0.5$, matched relative depth); the encoder is early-exited at layer 20, halving its forward cost. The predictor $\phi_\omega$ is a stack of 3 DiT blocks~\cite{peebles2023dit}, hidden dimension 1536, FFN ratio 4, 12 heads with QK-norm followed by a linear projection to dimension 5120. 
The predictor contains ${\sim}138$M parameters, accounting for ${\sim}9.7\%$ of the 1.42B parameters of Wan-1.3B, and is discarded during inference.

\paragraph{Training pipeline.}
We follow the three-stage Causal-Forcing pipeline~\cite{zhu2026causal}: Stage 1, teacher AR fine-tuning for 2K steps; Stage 2, ODE distillation for 2K steps; Stage 3, asymmetric DMD, which is run for 100, 500, and 600 steps for chunk-wise Causal-Forcing, frame-wise Causal-Forcing, and Self-Forcing, respectively, following each baseline's recipe.
Foresight is attached only during the final 100 steps of Stage 3 in our default configuration. We also discuss applying foresight in alternative stages in Appendix~\ref{sec:appendix:stages}.
At each training step, the causal generator rolls out a video, with mid-depth hidden states retained, and the foresight encoder processes the same full rollout. The foresight loss is computed on the clean rollout, while the asymmetric DMD loss is computed on a re-noised version, scored against a frozen Wan-14B real-score teacher and a Wan-1.3B fake-score critic. The fake-score critic is updated 5 times per 
generator update using the standard flow-matching loss, while the real-score model remains frozen throughout training.

\paragraph{Optimization.}
We use AdamW for the causal generator, predictor, and fake-score critic, with linear warmup. The causal generator and critic learning rates follow the
Causal-Forcing recipe; the predictor learning rate is $2 \times10^{-6}$. All distillation runs use $8\times$ H100 GPUs with FSDP sharding and gradient accumulation of 8, with an effective batch size of 64. Mixed-precision training uses bf16. Training prompts come from the filtered VidProM \cite{wang2024vidprom} extension released by Self-Forcing.

\paragraph{Evaluation.}
We evaluate on VBench~\cite{huang2024vbench}. For the 5-second generation, each metric is the mean over 5 generated videos per prompt across the full VBench prompt set, following the benchmark's standard protocol. The Total Score follows the standard VBench $0.8 \cdot$ Quality + $0.2 \cdot$ Semantic weighting. For the 30-second long-horizon evaluation, we roll out the same trained checkpoints on 200 randomly selected MovieGen prompts, following the Rolling-Forcing~\cite{liu2025rolling} protocol, and report the subset of quality and semantic dimensions that are computable under its prompt-agnostic setting, notably subject consistency, background consistency, and overall consistency, the dimensions most directly tied to causal short-sightedness.

\paragraph{Prompt overlap.}
\label{sec:appendix:prompt_overlap}
Training and evaluation prompts come from disjoint sources. We audited exact duplicates after Unicode, case, punctuation, and whitespace normalization between the 248,221 training prompts and both evaluation sets (946 VBench prompts, 1,003 MovieGen prompts) and found zero matches.

The pseudocode for one training step is given in Algorithm~\ref{alg:ff}.

\begin{algorithm}[t]
\caption{Video-Mirai training step}
\label{alg:ff}
\begin{algorithmic}[1]
\Require causal generator $G_\theta$, predictor $\phi_\omega$, frozen foresight
         encoder $E_\xi$, 
         fake-score critic $D_\psi$, frozen real-score model $D_{\text{real}}$,
         text prompt $c$, offset set
         $\Delta=\{0,\dots,K\}$, causal generator/encoder layers $L, L'$, weight $\lambda$

\State \Comment{Causal generator rollout with retained mid-depth hidden states}
\State Initialize ${\mathbf{x}} \gets \varnothing$, $\{\mathbf{h}_i^L\} \gets \varnothing$
\For{segment index $i = 1, \dots, N$}
    \State Denoise ${\mathbf{X}}_i$ conditioned on $c$ using $G_\theta$ with KV-cache of ${\mathbf{X}}_{<i}$  
    \State Cache $\mathbf{h}_i^L$ at layer $L$; append ${\mathbf{X}}_i$ to ${\mathbf{x}}$
\EndFor

\State \Comment{Foresight encoder: one forward over the full rollout}
\State $\{\mathbf{H}_j^{L'}\}_{j=1}^{N} \gets E_\xi({\mathbf{x}}, c)$ at layer $L'$,  early-exit.

\State \Comment{Foresight loss on valid segments}
\State $\bar{\mathbf{H}}_i \gets \frac{1}{|\Delta|}\sum_{\delta \in \Delta} \mathbf{H}_{i+\delta}^{L'}$ for $i = 1, \dots, N{-}K$
\State $\mathcal{L}_{\text{Foresight}} \gets \frac{1}{N-K}\sum_{i=1}^{N-K} \big[1 - \cos(\phi_\omega(\mathbf{h}_i^L),\, \text{sg}[\bar{\mathbf{H}}_i])\big]$

\State \Comment{Combined objective and update}
\State $\mathcal{L}_{\text{Generation}} \gets$ asymmetric DMD loss on re-noised rollout ${\mathbf{x_{noise}}}$, using frozen $D_{\text{real}}$ and $D_\psi$
\State $\mathcal{L}_{\text{total}} \gets \mathcal{L}_{\text{Generation}} + \lambda \cdot \mathcal{L}_{\text{Foresight}}$
\State Update $\theta$ and $\omega$ via $\nabla \mathcal{L}_{\text{total}}$
\State \Comment{Fake-score critic update (5 steps per generator step) and predictor replay}
\State Update $D_\psi$ for 5 steps on re-noised rollouts via flow-matching loss
\State Update $\omega$ on 4 of these 5 steps by replaying cached $(\mathbf{h}_i^L, \bar{\mathbf{H}}_i)$ pairs via $\mathcal{L}_{\text{Foresight}}$
\end{algorithmic}
\end{algorithm}

\section{Additional Computational Cost}
\label{sec:appendix:flops}

Video-Mirai introduces no inference-time overhead: the foresight encoder $E_\xi$ and predictor $\phi_\omega$ are discarded after training, leaving the trained model as a strictly causal AR decoder with unchanged FLOPs and KV-cache behavior. 
Training-time overhead has two components, which we account for below by FLOP estimation on the chunk-wise Causal-Forcing pipeline at our training resolution: latent shape $1{\times}21{\times}16{\times}60{\times}104$, patch size $1{\times}2{\times}2$, 
giving $M_f = (60/2) \times (104/2) = 1560$ tokens per frame and $M = 21 \times M_f = 32{,}760$ tokens per training sample, and report the measured wall-clock overhead.

\paragraph{Foresight encoder forward.}
Adding foresight to Stage~3, which is the DMD stage, requires one forward pass of encoder $E_\xi$ per generator update. The encoder we analyze here is the Wan-14B: $d = 5120$, FFN dim $13{,}824$, 40 heads, 40 layers. We early-exit at layer 20 of 40, matching the student's mid-depth layer 15 of 30, which halves the encoder's forward cost. 
Each Wan-14B layer at $M = 32{,}760$ tokens contributes approximately $8Md^2 + 4M^2 d + 4M d \cdot d_{\text{ffn}} \approx 38.1$~TFLOPs forward, so 20 layers add ${\sim}762$~TFLOPs per generator-update step. The encoder is frozen and stores no gradients, so its backward cost is zero.

\paragraph{Predictor forward and backward.}
The 3-block DiT predictor $\phi_\omega$ runs at the causal generator's hidden dimension $d_c = 1536$ with FFN ratio 4, FFN dim $6144$ and 12 attention heads. The predictor processes one segment of 4,680 tokens at a time. Each block contributes ${\sim}0.4$~TFLOPs per segment, so forward across 3 blocks and all segments is ${\sim}8.4$~TFLOPs and backward is ${\sim}17$~TFLOPs (${\sim}2{\times}$ forward). A final linear projection $\mathbb{R}^{1536} \to \mathbb{R}^{5120}$ maps $\phi_\omega$'s output into the encoder's hidden width for cosine alignment; its cost is ${\sim}1.5$~TFLOPs forward+backward at $M = 32{,}760$, bringing the predictor pass to ${\sim}27$~TFLOPs.

\paragraph{Parameter and optimizer-state overhead.}
The predictor $\phi_\omega$ contains ${\sim}138$M trainable parameters in total: ${\sim}130$M from the 3 DiT blocks plus ${\sim}7.9$M from the final $1536{\to}5120$ linear projection, which is ${\sim}9.7\%$ of the 1.42B-parameter Wan-1.3B causal generator. Under AdamW with FSDP sharding across our $8$ training GPUs, the per-GPU optimizer-state increase is approximately $207$MB $ \approx 138\text{M} \times 12$~bytes/param / $8$.

\paragraph{FLOP overhead.}
With foresight enabled, each generator update adds ${\sim}789$~TFLOPs ($762$ from $E_\xi$ and ${\sim}27$ from $\phi_\omega$). We additionally update $\phi_\omega$ on four of the five critic steps using cached hidden states, adding ${\sim}27$~TFLOPs per replay. 
This replay amortizes the encoder cost across the DMD cycle: $E_\xi$ runs only on the generator step, while the predictor benefits from more frequent updates. 

\paragraph{Empirical wall-clock measurement.}
We measure per-iteration training time on $8\times$ H100 GPUs over 100 training iterations with foresight enabled and disabled, under otherwise identical Stage~3 configuration. Each asymmetric DMD cycle spans five iterations: one runs a generator update followed by a critic update, and the other four run only a critic update, so we report the two iteration types separately in Table~\ref{tab:wallclock}. The measured per-cycle overhead is \textbf{14.3\%}.\looseness=1

\begin{table}[h]
\centering
\caption{\textbf{Empirical wall-clock training time per iteration.}
Measured on $8\times$ H100 over 100 training iterations each.
One cycle is one generator iteration plus four critic-only iterations.}
\label{tab:wallclock}
\setlength{\tabcolsep}{6pt}
\begin{tabular}{lccc}
\toprule
Iteration type & Baseline (s) & + Foresight (s) & Overhead \\
\midrule
Generator + critic   & $244$ & $292$ & $+20.0\%$ \\
Critic only          & $\phantom{0}86$ & $\phantom{0}95$ & $+10.6\%$ \\
Per-cycle ($1$ gen + $5$ critic) & $588$ & $672$ & $\mathbf{+14.3\%}$ \\
\bottomrule
\end{tabular}
\end{table}

\paragraph{Inference cost.}
At deployment, the student runs without $E_\xi$, $\phi_\omega$, or the foresight loss. The inference cost is identical to a standard causal Wan-1.3B forward pass, and the KV cache is maintained as in any causal AR generator. Real-time streaming properties like the latency per frame or throughput are also unchanged from the Causal-Forcing baseline.

\section{Limitations}
\label{sec:limitations}

Our foresight window is deliberately small: the default $\Delta = \{0, 1\}$ looks at most one chunk ahead, roughly 3 latent frames ($\sim$0.75 seconds at 16 fps). The target itself is computed by a bidirectional encoder that attends over the complete rollout. Our ablations show that naively extending the window further within the current recipe yields diminishing returns (Table~\ref{tab:foresight_window}): adding a two-step-ahead target ($\Delta = \{0, 1, 2\}$) improves Semantic but degrades Quality, and the number of valid (source, target) pairs shrinks as $K$ grows since supervising chunk $i$ requires access to chunk $i+K$. We hypothesize that the Quality drop at $\Delta = \{0, 1, 2\}$ arises because far-offset targets from a single stochastic rollout are noisy representatives of a multimodal future, and uniform averaging injects that variance into the target. This is consistent with Figure~\ref{fig:visualization_4_future}, where the readout resembles the average of multiple future rollouts. Distance-weighted fusion, multi-sample far targets, and learned fusion are natural next steps for building targets for multimodal futures. Given how clearly anticipation is internalized at short horizons (Figure~\ref{fig:quantitative_visualization}), an open question is whether a longer training-time foresight window could
deliver proportional gains at longer generation horizons. Promising directions include longer rollouts with position-reweighted foresight loss, hierarchical alignment at multiple timescales, and virtual chunks that extrapolate beyond the rollout horizon. We leave these to future work.

\section{Additional Results}

\subsection{Full VBench}

Tables~\ref{tab:vbench_quality} and \ref{tab:vbench_semantic} report the detailed per-dimension breakdown of VBench Quality and Semantic Scores, covering 7 and 9 dimensions respectively, for the Causal-Forcing baseline and after applying Video-Mirai.
We evaluate both 5-second generation under the VBench standard protocol and 30-second long-horizon generation using MovieGen prompts under the VBench prompt-agnostic protocol.

\begin{table*}[htbp]
  \centering
  \small
  \caption{\textbf{Per-dimension Quality Score on VBench.}}
  \vspace{1mm}
  \label{tab:vbench_quality}
  \begin{tabular}{lcccccccc}
    \toprule
    Method &
    \makecell{\scriptsize Subject\\\scriptsize Consistency} &
    \makecell{\scriptsize Background\\\scriptsize Consistency} &
    \makecell{\scriptsize Temporal\\\scriptsize Flickering} &
    \makecell{\scriptsize Motion\\\scriptsize Smoothness} &
    \makecell{\scriptsize Dynamic\\\scriptsize Degree} &
    \makecell{\scriptsize Aesthetic\\\scriptsize Quality} &
    \makecell{\scriptsize Imaging\\\scriptsize Quality} &
    \makecell{\scriptsize Quality\\\scriptsize Score} \\
    \midrule
    \rowcolor{gray!20}
    \multicolumn{9}{l}{\emph{5-second}} \\
    Causal-Forcing & 96.05 & 95.92 & 98.64 & 98.26 & 61.11 & 67.01 & 70.92 & 84.54 \\
    + Video-Mirai    & \textbf{96.41} & \textbf{96.54} & \textbf{99.42} & \textbf{98.33} & \textbf{65.00} & \textbf{67.88} & 69.87 & \textbf{85.38} \\
     \midrule
    \rowcolor{gray!20}
    \multicolumn{9}{l}{\emph{30-second}} \\
    Causal-Forcing & 84.93 & 90.22 & 96.04 & 97.21 & 49.00 & 56.90 & 64.54 & 76.26 \\
    + Video-Mirai    & \textbf{88.47} & \textbf{91.94} & 95.93 & 97.16 & \textbf{49.50} & \textbf{61.52} & 64.25 & \textbf{77.88} \\
    \bottomrule
  \end{tabular}
\end{table*}
\begin{table*}[htbp]
  \centering
  \small
  \caption{\textbf{Per-dimension Semantic Score on VBench.} The 30-second evaluation provides only Temporal Style and Overall Consistency, which use the same ViCLIP video-text similarity on the same prompts and therefore coincide.}
  \vspace{1mm}
  \label{tab:vbench_semantic}
  \setlength{\tabcolsep}{4pt}
  \begin{tabular}{lcccccccccc}
    \toprule
    Method &
    \makecell{\scriptsize Object\\\scriptsize Class} &
    \makecell{\scriptsize Multiple\\\scriptsize Objects} &
    \makecell{\scriptsize Human\\\scriptsize Action} &
    {\scriptsize Color} &
    \makecell{\scriptsize Spatial\\\scriptsize Relation} &
    {\scriptsize Scene} &
    \makecell{\scriptsize Temporal\\\scriptsize Style} &
    \makecell{\scriptsize Appearance\\\scriptsize Style} &
    \makecell{\scriptsize Overall\\\scriptsize Consistency} &
    \makecell{\scriptsize Semantic\\\scriptsize Score} \\
    \midrule
    \rowcolor{gray!20}
    \multicolumn{11}{l}{\emph{5-second}} \\
    Causal-Forcing & 95.98 & 88.66 & 95.60 & 86.89 & 78.40 & 57.05 & 24.62 & 20.63 & 26.83 & 80.93 \\
    + Video-Mirai    & \textbf{96.79} & 87.96 & 95.40 & \textbf{87.26} & \textbf{82.29} & \textbf{57.57} & \textbf{24.83} & \textbf{20.76} & \textbf{26.85} & \textbf{81.59} \\
    \midrule
    \rowcolor{gray!20}
    \multicolumn{11}{l}{\emph{30-second}} \\
    Causal-Forcing & -- & -- & -- & -- & -- & -- & 24.93 & -- & 24.93 & -- \\
    + Video-Mirai    & -- & -- & -- & -- & -- & -- & \textbf{25.03} & -- & \textbf{25.03} & -- \\
    \bottomrule
  \end{tabular}
\end{table*}

\newpage
\subsection{Frame-wise Foresight Window}

\begin{table}[htbp]
\centering
\caption{\textbf{Frame-wise foresight window comparison.} Effect of the
offset set $\Delta$ when applied to the frame-wise Causal-Forcing model.}
\label{tab:framewise_foresight_window}
\begin{tabular}{lcccc}
\toprule
Setting & $\Delta$ & Quality $\uparrow$ & Semantic $\uparrow$ & Total $\uparrow$ \\
\midrule
Causal-Forcing (frame-wise) & -- & 83.16 & 78.73 & 82.27 \\
\midrule
Current only & $\{0\}$ & 83.81 & 79.50 & 82.94 \\
1-frame ahead only & $\{1\}$ & 83.79 & \textbf{79.66} & 82.96 \\
Current + 1-frame ahead & $\{0, 1\}$ & \textbf{84.59} & \textbf{79.66} & \textbf{83.60} \\
Current + 2-frame ahead & $\{0, 1, 2\}$ & 83.25 & 77.80 & 82.16 \\
\bottomrule
\end{tabular}
\end{table}

Table~\ref{tab:framewise_foresight_window} extends our foresight window study (Table~\ref{tab:foresight_window}) to the frame-wise setting, where each segment is a single latent frame. 
The chunk-wise conclusion holds: $\Delta{=}\{0,1\}$ achieves the best Total Score. Unlike the chunk-wise setting, extending to $\{0,1,2\}$ drops the Total Score below the baseline.

\subsection{Loss Weight}
\label{sec:appendix:loss_weight}
Table~\ref{tab:loss_weight} sweeps the foresight weight $\lambda \in \{0.1, 0.2, 0.3\}$ on Causal-Forcing.
$\lambda{=}0.2$ achieves the best Total Score: lower weights under-supervise the predictor and yield smaller gains, while higher weights over-pull the causal generator's mid-depth features toward the encoder and slightly degrade the Quality Score. We adopt $\lambda{=}0.2$ throughout.
\begin{table}[htbp]
  \centering
    \caption{\textbf{Foresight weight $\lambda$ ablation} on
  Causal-Forcing.}
  \label{tab:loss_weight}
  \begin{tabular}{lccc}
    \toprule
      Weight & Quality $\uparrow$ & Semantic $\uparrow$ & Total $\uparrow$ \\ 
    \midrule
        Causal-Forcing & 84.54 & 80.93 & 83.82 \\
      \midrule
    0.1  & 84.83 & 81.42 & 84.15\\
    0.2 &  \textbf{85.38} & 81.59 & \textbf{84.62} \\
    0.3  & 85.14 & \textbf{81.71} & 84.45  \\
    \bottomrule
  \end{tabular}
    \vspace{-2mm}
\end{table}

\subsection{Foresight at Different Distillation Stages}
\label{sec:appendix:stages}

Causal-Forcing exposes two training stages at which Video-Mirai can be attached: Stage~2, causal ODE distillation, and Stage~3, asymmetric DMD. These two stages are not interchangeable injection points because they differ in what data the causal generator consumes. In this section, we analyze how to inject foresight into Stage 2.

\paragraph{Stage~2 ODE: target from the ground-truth future.}
Stage~2 trains the causal generator's flow map on paired $(\text{noise} \to \text{clean})$ trajectories whose clean side is a ground-truth video $\{\mathbf{X}_1^{\text{GT}}, \dots, \mathbf{X}_N^{\text{GT}}\}$. The causal generator is teacher-forced on clean GT prefixes, and the ODE distillation loss is defined on GT trajectories. The consistent foresight target is therefore the GT future segment: at each step, we feed the clean latent of
$\mathbf{X}_{i+\delta}^{\text{GT}}$ to the foresight encoder. Both the ODE loss and the foresight loss live on the same clean, teacher-forced manifold.

\paragraph{Stage~3 DMD: target from the causal generator's own rollout.}
Stage~3 removes the paired data: the causal generator self-rolls a video $\{{\mathbf{X}}_1, \dots, {\mathbf{X}}_N\}$ from noise, and asymmetric DMD matches the causal generator's output distribution to a bidirectional teacher's score. The consistent foresight target is the foresight encoder's representation of the causal generator's own future rollout. Foresight then regularizes the same trajectory and distribution that DMD already matches, ensuring the two objectives operate on a shared signal.

\paragraph{Ablation study of foresight in Stage~2.}
\label{sec:appendix:stage2_encoder}

The main paper studies foresight at Stage~3 (DMD). For completeness, we conduct two parallel ablations within Stage~2 (ODE) along the same axes as the main paper's Stage~3 counterparts: the foresight encoder and the predictor's fusion strategy. All Stage~2 ablation runs use a shortened budget of 1K distillation steps and no Stage~3 refinement.

Table~\ref{tab:stage2_preditcor} fixes the encoder to Wan-14B and sweeps three fusion strategies for the offset set $\Delta{=}\{0,1\}$: no fusion (one loss per offset), per-frame averaging (encoder features averaged across frames within each segment), and per-$\delta$ averaging (encoder features averaged across offsets, i.e., the default in the main paper, Eq.~\ref{eq:foresight}). Table~\ref{tab:stage2_encoder} fixes the fusion strategy to per-frame averaging and sweeps the encoder among EMA, Wan-1.3B, and Wan-14B. 
Two findings emerge. First, at Stage~2, per-$\delta$ averaging is actively harmful, while per-frame averaging is best. This contrasts with Stage~3, where per-$\delta$ target fusion is optimal. The two stages train on different data: ODE pairs of GT videos versus the causal generator's own rollouts. The fusion strategy that aligns with each stage's data is what wins. Second, Wan-14B remains the strongest encoder at Stage~2, as at Stage~3 (Table~\ref{tab:foresight_encoder}).

\begin{table}[htbp]
  \centering
  \caption{\textbf{Stage~2 fusion-strategy ablation.} 
  Encoder fixed to Wan-14B after 1K training steps.}
  \label{tab:stage2_preditcor}
  \begin{tabular}{lccc}
    \toprule
     Fusion strategy & Quality $\uparrow$ & Semantic $\uparrow$ & Total $\uparrow$ \\
    \midrule
    Causal-Forcing (stage 2 reproduction)  &  82.56 & \textbf{78.30} & 81.71 \\
      \midrule
    no fusion & 82.65 & 77.51 & 81.62   \\
    per-frame averaging & \textbf{84.17} & \textbf{77.68} & \textbf{82.87}  \\
    per-$\delta$ averaging  & 81.45 & 74.32 & 80.02  \\
    \bottomrule
  \end{tabular}
\end{table}
\begin{table}[htbp]
  \centering
  \caption{\textbf{Stage~2 encoder ablation.} 
  Fusion strategy fixed to per-frame averaging after 1K training steps.}
  \label{tab:stage2_encoder}
  \begin{tabular}{lccc}
    \toprule
    Encoder & Quality $\uparrow$ & Semantic $\uparrow$ & Total $\uparrow$ \\
    \midrule
    Causal-Forcing (stage 2 reproduction)  &  82.56 & 78.30 & 81.71 \\
      \midrule
    EMA & 83.96 & 78.01 & 82.77 \\
    Wan-1.3B & 83.40 & \textbf{78.76} & 82.47  \\
    Wan-14B & \textbf{84.17} & 77.68 & \textbf{82.87}  \\
    \bottomrule
  \end{tabular}
\end{table}

\paragraph{Foresight injection stage comparison.}
Causal-Forcing's three-stage pipeline exposes two natural injection points for foresight: ODE distillation (Stage~2) and asymmetric DMD (Stage~3). Whether to attach foresight at Stage~2, Stage~3, or both is not \emph{a priori} obvious. Table~\ref{tab:foresight_stage} sweeps all three options.
The results show that ODE-Foresight and DMD-Foresight yield identical total gains over Causal-Forcing, but split the credit very differently: ODE-Foresight drives the largest quality gain while slightly reducing the Semantic Score, consistent with pulling the flow map toward a clean, well-formed future, whereas 
DMD-Foresight drives the largest semantic gain while retaining most of the quality improvement, consistent with regularizing the very distribution DMD is already matching, on the same rollout trajectory.

Stacking the two does not combine these profiles, collapsing to a Total Score only marginally above the baseline. 
The reason is that the two instantiations do not disagree about \emph{whether} the causal generator should anticipate the future, but about \emph{which} future: 
ODE-Foresight trains $\phi_\omega$ to predict a clean GT future from a GT prefix, while DMD-Foresight trains it to predict the future from the causal generator's \emph{own} rollout, and the two trajectories differ whenever the causal generator is imperfect. Stacked naively, Stage~3 must rewrite the foresight mapping Stage~2 just installed, swapping a clean target for a moving one while simultaneously running asymmetric DMD; these pressures pull mid-depth features in conflicting directions.

DMD-Foresight is the most practical choice since it is equally effective and cheaper: it drops into any pretrained Causal-Forcing Stage~2 checkpoint without modifying Stage~2 training, and requires only 100 DMD steps, 20$\times$ fewer than the 2K Stage~2 distillation steps. We therefore use it throughout the main paper.

\begin{table}[htbp]
  \centering
   \caption{\textbf{Foresight injection stage comparison.} Foresight applied 
   at Stage~2 (ODE-Foresight), Stage~3 (DMD-Foresight), or both. }
  \label{tab:foresight_stage}
  \begin{tabular}{lccc}
    \toprule
     Stage & Quality $\uparrow$ & Semantic $\uparrow$ & Total $\uparrow$ \\
    \midrule
    Causal-Forcing   & 84.54 & 80.93 & 83.82 \\
      \midrule
    Stage 2  &  \textbf{85.56} & 80.84 & \textbf{84.62} \\
    Stage 3 &  85.38 & \textbf{81.59} & \textbf{84.62} \\
    Stage 2 + 3  & 84.78 & 81.10 & 84.05  \\
    \bottomrule
  \end{tabular}
\end{table}

\begin{figure*}[t]
  \centering
    \includegraphics[width=\linewidth]{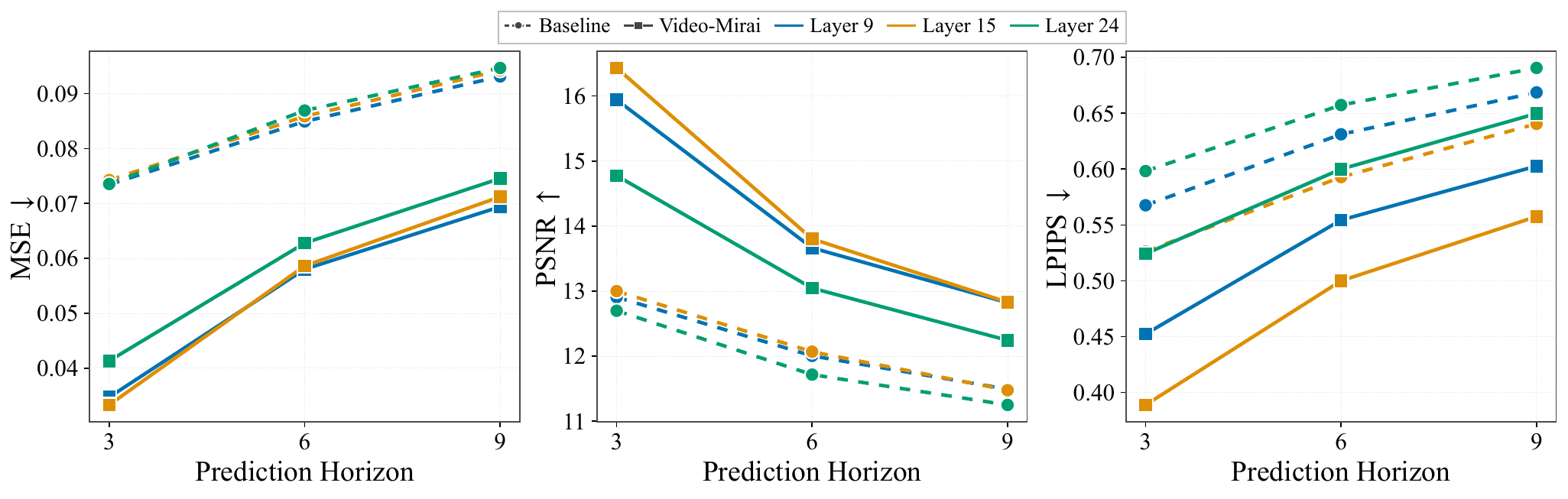}
      \vspace{-6mm}
\caption{\textbf{Future-frame readout fidelity across layers and horizons.} Solid: Video-Mirai; dashed: Causal-Forcing baseline. Video-Mirai dominates at every layer and horizon.}
    \label{fig:quantitative_visualization_aapendix}
\end{figure*}

\subsection{Foresight Encoder from Outside the Model Family}
\label{sec:appendix:dino}
To test whether the foresight encoder must come from the same model family as the generator, we replace Wan-14B with a frozen 21M-parameter DINO ViT-S/16~\cite{caron2021dino}. DINO is a self-supervised image encoder that is neither a diffusion model nor trained on video, and it receives no text conditioning. Its patch grid matches the causal generator's latent token grid one-to-one, so the rest of the recipe is unchanged: offsets $\{0,1\}$ and the same 100 DMD steps. Table~\ref{tab:dino_encoder} shows that this much smaller encoder recovers about two thirds of the gain of Wan-14B, so the foresight encoder need not be a Wan model or a video diffusion model. We keep Wan-14B as the default because it is already resident in the pipeline as the DMD real-score teacher and shares the VAE latent space.

\begin{table}[htbp]
  \centering
  \caption{\textbf{Foresight encoder from outside the model family.} Causal-Forcing chunk-wise on 5-second VBench; DINO ViT-S/16 replaces Wan-14B with the rest of the recipe unchanged.}
  \label{tab:dino_encoder}
  \begin{tabular}{lccc}
    \toprule
    Encoder & Quality $\uparrow$ & Semantic $\uparrow$ & Total $\uparrow$ \\
    \midrule
    Causal-Forcing & 84.54 & 80.93 & 83.82 \\
    \midrule
    Wan-14B & \textbf{85.38} & \textbf{81.59} & \textbf{84.62} \\
    DINO ViT-S/16 & 85.09 & 81.33 & 84.34 \\
    \bottomrule
  \end{tabular}
\end{table}

\subsection{60-Second Generation}
\label{sec:appendix:movie60s}
We extend the same checkpoints to 60-second rollouts on the same 200 MovieGen prompts and seeds as the 30-second evaluation, under the same prompt-agnostic protocol. Comparing Table~\ref{tab:movie60s} with the 5- and 30-second results in Tables~\ref{tab:system_comparison_consistency} and~\ref{tab:vbench_quality} shows that the gains increase beyond the training horizon. From 5 to 30 to 60 seconds, the gain grows from $+0.36$ to $+3.54$ to $+10.35$ on Subject Consistency, from $+0.62$ to $+1.72$ to $+4.51$ on Background Consistency, from $+0.02$ to $+0.10$ to $+0.77$ on Overall Consistency, and from $+0.84$ to $+1.62$ to $+7.49$ on the seven-dimension Quality Score.

\begin{table}[htbp]
  \centering
  \caption{\textbf{60-second generation.} 200 MovieGen prompts under the prompt-agnostic protocol, 969 decoded frames per video.}
  \label{tab:movie60s}
  \setlength{\tabcolsep}{10pt}
  \begin{tabular}{lcccc}
    \toprule
    Method & \makecell{Subject\\Consistency} $\uparrow$ & \makecell{Background\\Consistency} $\uparrow$ & \makecell{Overall\\Consistency} $\uparrow$ & \makecell{Quality\\Score} $\uparrow$ \\
    \midrule
    Causal-Forcing & 73.74 & 85.03 & 23.32 & 65.81 \\
    + Video-Mirai & \textbf{84.09} & \textbf{89.54} & \textbf{24.09} & \textbf{73.30} \\
    \bottomrule
  \end{tabular}
\end{table}

\section{Statistical Significance}
\label{sec:appendix:stats}

We test whether Video-Mirai's gains over Causal-Forcing chunk-wise are statistically significant via a paired prompt-level bootstrap. 
For each VBench dimension, we resample its $N_d$ prompts with replacement, using the same indices for both methods and recompute the dimension's mean, repeating 10{,}000 times. Each replicate is then aggregated into Quality, Semantic, and Total Scores using VBench's standard $\tfrac{4}{5}\,\text{quality} + \tfrac{1}{5}\,\text{semantic}$ weighting~\cite{huang2024vbench}, yielding bootstrap distributions of the three summary metrics. We report the mean paired difference with its standard error; a $^{*}$ marks a $[2.5\%, 97.5\%]$ percentile $95\%$ CI that excludes zero (two-sided $p < 0.05$). Per-prompt scores are obtained by averaging the 5 inference seeds per prompt.

Table~\ref{tab:bootstrap_ci} shows that Video-Mirai's gains on all three aggregated metrics have $95\%$ CIs strictly excluding zero, confirming the improvements are not driven by a small subset of favorable prompts.

\begin{table}[h]
\centering
\caption{\textbf{Significance of Video-Mirai gains over 
Causal-Forcing on VBench.} $10{,}000$ paired bootstrap resamples by prompt; cells: $\Delta_{\text{mean}} \pm \text{SE}$. $^{*}$ marks $95\%$ confidence intervals excluding zero, equivalent to a two-sided $p < 0.05$ test.}
\label{tab:bootstrap_ci}
\setlength{\tabcolsep}{8pt}
\begin{tabular}{lccc}
\toprule
Method  & $\Delta$Quality & $\Delta$Semantic & $\Delta$Total \\
\midrule
Video-Mirai (vs Causal-Forcing) & $+0.84 \pm 0.19^{*}$ & $+0.66 \pm 0.24^{*}$ & $+0.80 \pm 0.15^{*}$ \\
\bottomrule
\end{tabular}
\end{table}

\section{More Probing Results}
\label{sec:appendix:probing}
\begin{figure*}[h]
  \centering
    \includegraphics[width=\linewidth]{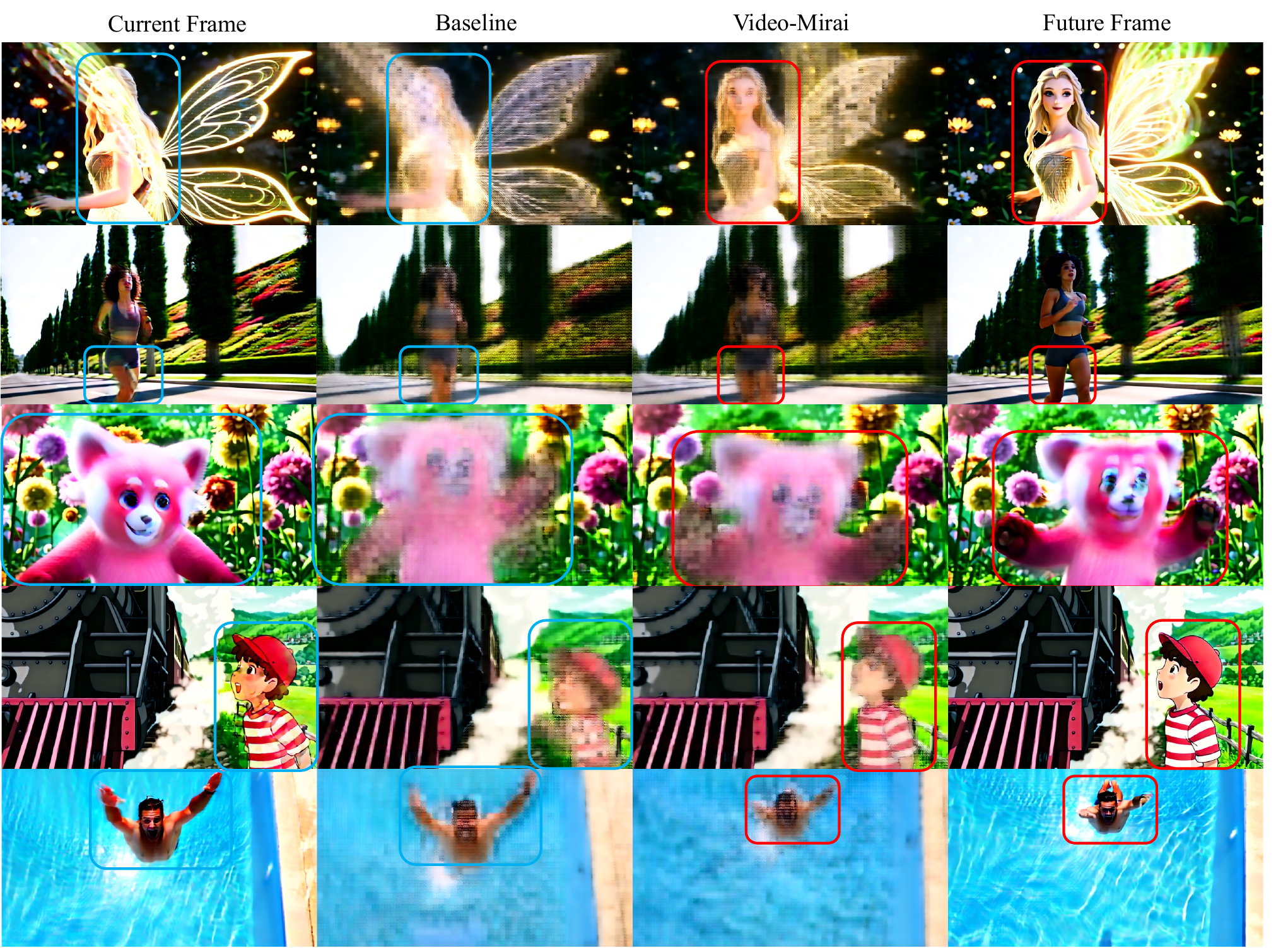}
    \vspace{-0.2in}
    \caption{\textbf{Video-Mirai's representations encode the future.} An MLP readout reconstructs future RGB from the frozen causal generator’s current hidden state (layer 15). Left to right: current frame, baseline readout, Video-Mirai readout (ours), future frame. \textcolor{colorcurrent}{blue}/\textcolor{colorfuture}{red}: regions matching the \textcolor{colorcurrent}{current}/\textcolor{colorfuture}{future} frame.}
\label{fig:visualization_appendix}
\vspace{-0.1in}
\end{figure*}

We provide additional visualization results of our probing analysis in Figures \ref{fig:quantitative_visualization_aapendix} and \ref{fig:visualization_appendix}.

\section{More Comparison Results}
\label{sec:appendix:comparison_results}
\begin{figure*}[h]
  \centering
    \includegraphics[width=\linewidth]{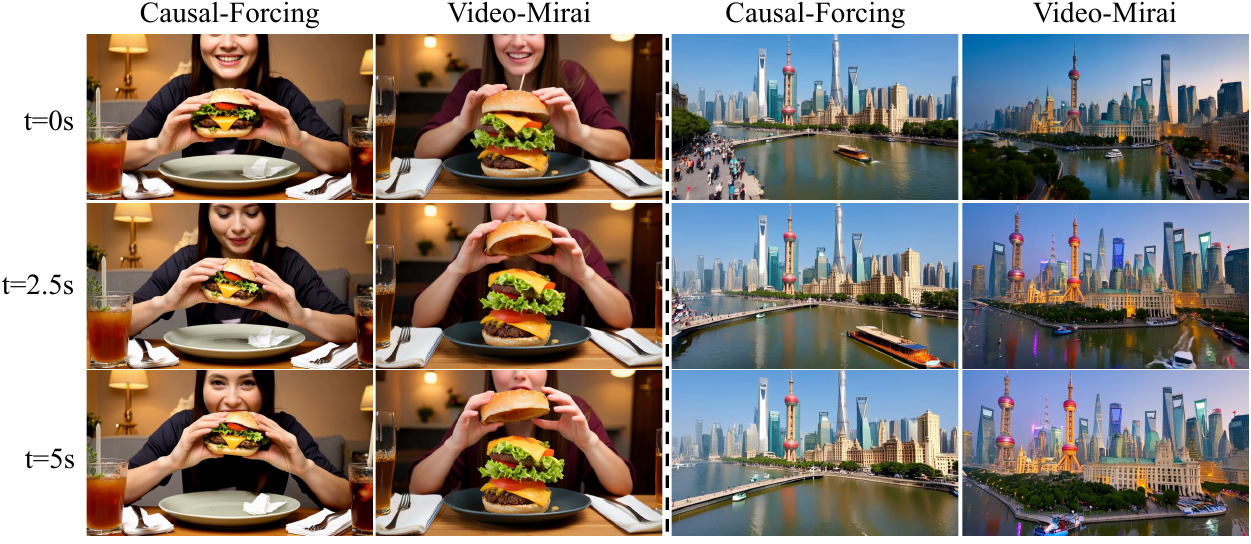}
    \vspace{-5mm}
    \caption{\textbf{More qualitative comparison.} Representative frames at $t{=}0$s, $2.5$s, $5$s from the rollouts of the same prompt under the baseline and its Video-Mirai counterpart. Video-Mirai rollouts preserve subject identity, scene composition, and motion coherence more reliably across time.}
    \label{fig:video_sample_appendix}
    \vspace{-4mm}
\end{figure*}

We provide additional visual comparisons in Figure \ref{fig:video_sample_appendix}.

\end{document}